\documentclass[smallextended]{svjour3}

\usepackage{amsmath,bm}
\usepackage{amsthm}
\usepackage{amsfonts}
\usepackage{amssymb}
\usepackage{mathtools}
\usepackage{graphicx}
\usepackage{subfigure}
\usepackage{xspace}
\usepackage{url}
\usepackage{longtable}
\usepackage{booktabs}
\usepackage{threeparttable}
\usepackage[pdftex,dvipsnames]{xcolor}
\usepackage[lined,boxed,commentsnumbered,ruled,linesnumbered]{algorithm2e}
\usepackage{xargs} 
\usepackage{hyperref}
\usepackage[open,openlevel=3]{bookmark}
\usepackage{placeins} 
\usepackage{afterpage}

\usepackage[square,sort,comma,numbers]{natbib}

\newcommand\myshade{90}
\colorlet{mylinkcolor}{violet}
\colorlet{mycitecolor}{YellowOrange}
\colorlet{myurlcolor}{Aquamarine}

\hypersetup{
    linkcolor = mylinkcolor!\myshade!black,
    citecolor = mycitecolor!\myshade!black,
    urlcolor = myurlcolor!\myshade!black,
    colorlinks = true,
}

\usepackage{etoolbox}
\newcommand*{\affaddr}[1]{#1} 
\newcommand*{\affmark}[1][*]{\textsuperscript{#1}}

\begin{document}

    \newcommand\ie[0]{\textit{i.e.} }
    \newcommand\eg[0]{\textit{e.g.} }

    \newcommand{\ourmethod}{Multivariate Elastic Measures\xspace}
    \newcommand{\ourmethodlong}{Multivariate Elastic Similarity \reviewtwo{and Distance} Measures}
    \newcommand{\gitrepo}{\url{https://github.com/dotnet54/multivariate-measures}}


\definecolor{shcolor}{rgb}{0.758, 0.188, 0.478}
\newcommand{\sh}[1]{\textcolor{shcolor}{#1}}

    \newcommand{\cpc}[1]{\textcolor{magenta}{\textbf{CH}: #1}}

    \newcommand{\gwc}[1]{\textcolor{green}{#1}}

    \newcommand{\reviewone}[1]{{\color{black}{}#1}}
    \newcommand{\reviewtwo}[1]{{\color{black}{}#1}}
    \newcommand{\reviewthree}[1]{{\color{black}{}#1}}

\sloppy





    \title{Elastic Similarity \reviewtwo{and Distance} Measures for Multivariate Time Series
    \thanks{This research has been supported by Australian Research Council grant DP210100072.}}
    \titlerunning{Multivariate Elastic Similarity \reviewtwo{and Distance} Measures} 
    \authorrunning{Shifaz et al.}

    \author{Ahmed~Shifaz\affmark[1]\and Charlotte~Pelletier\affmark[1]\textsuperscript{,}\affmark[2] \and
        Fran\c{c}ois~Petitjean\affmark[1] \and Geoffrey~I.~Webb\affmark[1]}

    \institute{
        \affaddr{\affmark[1]Department of Data Science and Artifical Intelligence}\\
        Monash University, Melbourne \\
        VIC 3800, Australia\\
        \affaddr{\affmark[2] IRISA, UMR CNRS 6074}\\
        Univ. Bretagne Sud\\
        Campus de Tohannic\\
        BP 573, 56 000 Vannes, France\\
        \email{ahmedshifaz.mv@gmail.com, \\
            \{ahmed.shifaz,francois.petitjean,geoff.webb\}@monash.edu,\\
            charlotte.pelletier@univ-ubs.fr}
    }

    \date{Received: 27 June 2021 / Revised: 9 November 2022 / Accepted: 3 December 2022}


        
    
    
    
    
    
    

\maketitle

\begin{abstract}

This paper contributes multivariate versions of seven commonly used elastic similarity \reviewtwo{and distance} measures for time series data analytics. Elastic similarity \reviewtwo{and distance} measures can compensate for misalignments in the time axis of time series data. We adapt two existing strategies used in a multivariate version of the well-known Dynamic Time Warping (DTW), namely, Independent and Dependent DTW, to these seven measures.

While these measures can be applied to various time series analysis tasks, we demonstrate their utility on multivariate time series classification using the nearest neighbor classifier. On 23 well-known datasets, we demonstrate that \reviewone{each of the measures but one achieve the highest} accuracy relative to others on at least one dataset, \reviewone{supporting the value} of developing a suite of multivariate similarity \reviewtwo{and distance} measures. We also demonstrate that there are datasets for which either the dependent versions of all measures are more accurate than their independent counterparts or vice versa. 
\reviewone{In addition, we also construct a nearest neighbor based ensemble of the measures and show that it is competitive to other  state-of-the-art single-strategy multivariate time series classifiers.}


\end{abstract}




\section{Introduction}
\label{sec:intro}

Elastic similarity \reviewtwo{and distance} measures, such as the well known Dynamic Time Warping (DTW)~\cite{sakoe1978dynamic}, are a key tool in many forms of time series analytics. Elastic similarity \reviewtwo{and distance} measures can align temporal misalignments between two series while computing the similarity or distance between them.
Examples of their application include  clustering~\cite{berndt1994using, aghabozorgi2015time, liao2005clustering}, classification~\cite{lines2015time, bagnall2017great}, anomaly detection~\cite{izakian2014anomaly, steiger2014visual}, indexing~\cite{gunopulos2001time}, subsequence search~\cite{park2001segment} and segmentation~\cite{cassisi2012similarity}.

While numerous elastic similarity \reviewtwo{and distance} measures have been developed~\cite{ding2008querying, bagnall2017great, keogh2003need}, most of these measures have been defined only for univariate time series. 
One elastic measure that has previously been extended to the multivariate case is DTW~\cite{shokoohi2017generalizing}. That work identified two key strategies for such extension.
The \emph{independent} strategy applies the univariate measure to each dimension and then sums
the resulting distances. The \emph{dependent} strategy treats the multivariate series as a single series in which each time step has a single multi-dimensional point.
DTW is then applied using Euclidean distances between the multidimensional points of the two series.
%
%

This paper extends seven further key univariate similarity \reviewtwo{and distance} measures, \reviewone{presented in Table~\ref{tbl:measures},} to the multivariate case. 
We choose these seven specific measures because our research is largely focused on classification and these measures have been used in many well known univariate similarity \reviewtwo{and distance}-based classifiers such as Elastic Ensemble~\cite{lines2015time}, Proximity Forest~\cite{lucas2019proximity}, and two state-of-the-art univariate time series classifiers HIVE-COTE 1.0~\cite{lines2018time} and TS-CHIEF~\cite{shifaz2020ts}. 
\reviewone{

\begin{table}[ht]
\centering
\caption{Measures used in this paper. Measure name with subscript $I$ indicates independent measures and subscript $D$ indicates dependent measures. $L$ is the length of the time series and $D$ is the number of dimensions. The seven measures we extend to the multivariate case in this paper are: DDTW, WDTW, WDDTW, LCSS, ERP, MSM and TWE.}
\label{tbl:measures}
\begin{tabular}{rlll} 
\toprule
                    Measure & Parameters & Time Complexity & Reference \\
\midrule
$Lp$ (e.g. Euclidean) & N/A & $O(L \cdot D)$ & N/A \\ [0.5em]
$\mathit{DTWF_I}$, $DTWF_D$ & N/A (full window) & $O(L^2 \cdot D)$ & \cite{sakoe1978dynamic, shokoohi2017generalizing} \\ [0.5em]
$DTW_I$, $DTW_D$ & window size $w$ & $O(L \cdot w \cdot D)$ & \cite{sakoe1978dynamic} \\ [0.5em]
$\mathit{DDTWF_I}$, $\mathit{DDTWF_D}$ & N/A (full window) & $O(L^2 \cdot D)$ & \cite{keogh2001derivative} \\ [0.5em]
$DDTW_I$, $DDTW_D$ & window size $w$ & $O(L \cdot w \cdot D)$ & \cite{keogh2001derivative} \\ [0.5em]
$\mathit{WDTW_I}$, $\mathit{WDTW_D}$ & N/A & $O(L^2 \cdot D)$ & \cite{jeong2011weighted} \\ [0.5em]
$\mathit{WDDTW_I}$, $WDDTW_D$ & N/A & $O(L^2 \cdot D)$ & \cite{lines2015time} \\ [0.5em]
$LCSS_I$, $\mathit{LCSS_D}$ & window size $w,\epsilon $ & $O(L \cdot w \cdot D)$ & \cite{hirschberg1977algorithms, vlachos2002discovering} \\ [0.5em]
$ERP_I$, $ERP_D$ & window size $w$, penalty $g$ & $O(L \cdot w \cdot D)$ & \cite{chen2004marriage, chen2005robust} \\ [0.5em]
$MSM_I$, $MSM_D$ & cost $c$ & $O(L^2 \cdot D)$ & \cite{stefan2012move} \\  [0.5em]
$TWE_I$, $TWE_D$ & stiffness $\nu$, penalty $\lambda$ & $O(L^2 \cdot D)$ & ~\cite{marteau2008time} \\ [0.5em]
\bottomrule
\end{tabular}
\end{table}
}

This extension is important, because many real-world time series are multi-dimensional. For some measures it is straightforward, but non-trivial for \reviewone{three measures, LCSS, MSM, and TWE}. 
We show that each measure except one provides more accurate nearest neighbor classification than any alternative for at least one dataset. This demonstrates the importance of having a range of multivariate elastic measures.


It has been shown that the dependent and independent strategies each outperformed the other on some tasks when applied to DTW~\cite{shokoohi2017generalizing}.
One of this paper's significant outcomes is to demonstrate that there are some tasks for which the independent strategy is superior across all measures and others for which the dependent strategy is better.
This establishes a fundamental relationship between the two strategies and different tasks, countering the possibility that differing performance for the two strategies when applied to DTW might have been
coincidental.

We further illustrate the value of multiple measures by developing a multivariate version of the Elastic Ensemble~\cite{lines2015time}. We demonstrate that this ensemble of nearest neighbor classifiers using all multivariate measures provides accuracy competitive with the state-of-the-art single strategies in multivariate time series classification.


We organize the rest of the paper as follows. Section~\ref{sec:background} presents key definitions and  a brief
review of existing methods.
Section~\ref{sec:measures} describes our new multivariate similarity \reviewtwo{and distance} measures.
Section~\ref{sec:experiments} presents multivariate time series classification experiments on the UEA
multivariate time series archive, and includes
discussion of the implications of the results.
Finally, we draw conclusions in Section~\ref{sec:conclusion}, with suggestions for future work.


\vspace*{-10pt}
\section{Related Work}
\label{sec:background}



\subsection{Definitions}
\label{sec:definitions}
We here present key notations and definitions.


    A \emph{time series} $T$ of length $L$ is an ordered sequence of $L$ time-value pairs
    $ T = \langle (t_1, \bm{x}_1), \cdots, (t_L, \bm{x}_L) \rangle$, where $t_i$
    is the timestamp at sequence index $i$, $i \in \{1, \cdots, L\}$, and $\bm{x}_i$ is a $D$-dimensional
    point representing observations of $D$ real-valued variables or features at timestamp $t_i$.
    Each time point $\bm{x}_i\in \mathbb{R}^D$ is defined by $\{x^1_i, \cdots ,x^d_i, \cdots ,x^D_i \}$.
    Usually, timestamps $t_i$ are assumed to be equidistant, and thus omitted, which results in a simpler
    representation where~$T~=~\langle \bm{x}_1,\cdots, \bm{x}_L \rangle$.

    A \emph{univariate} (or single-dimensional) time series is a special case where a single variable is
    observed ($D=1$). Therefore, $\bm{x}_i$ is a scalar, and consequently,
    $T = \langle x_1, \cdots, x_L\rangle$.




    A \emph{labeled time series datase}t $\mathcal{S}$ consists of $n$ labeled time series indexed by $k$,
    where $k \in \{1, \cdots, n\}$.
    Each time series $T_k$ in $\mathcal{S}$ is associated with a label $y_k\in\{1,\cdots,c\}$, where $c$ is the number of classes. 

\label{dfn:similarity-measure}

A \emph{similarity measure} computes a real value that quantifies similarity
between two sets of values.  A \emph{distance measure} computes a real value that quantifies dissimilarity
between two sets of values.
For time series $Q$ and $C$, a similarity or distance measure $M$ is defined as

\begin{equation}
    M(Q, C) \rightarrow \mathbb{R}
    \label{eq:measure}
\end{equation}
A measure $M$ is a \emph{metric} if it has the following properties:
\reviewone{
\begin{enumerate}
    \item Non-negativity: $M(Q, C) \geq 0$,
    \item Identity: $M(Q, C) = 0$, if and only if $Q = C$,
    
    \item Symmetry: $M(Q, C) = M(C, Q)$,
    \item Triangle Inequality: $M(Q, C) \leq M(Q, T) + M(T, C)$ for any time series $Q, C$ and $T$.
\end{enumerate}
}


    In a Time Series Classification (TSC) task, a time series classifier is trained on a labeled time series dataset, 
    and then used to predict labels of unlabeled time series.
    The classifier is a predictive mapping function that maps from the space of input variables
    to discrete class labels.

    In this paper, to perform TSC tasks, we use 1-nearest neighbor (1-NN) classifiers, which use time series specific {similarity \reviewtwo{and distance} measures} to compute the nearest neighbors between each time series.



\subsection{Univariate TSC }\label{subsec:uni-tsc}
A comprehensive review of the most common univariate TSC methods developed prior to 2017 can be found in \cite{bagnall2017great}. Here we summarize key univariate TSC methods following a widely used categorization as follows:

\begin{itemize}
    \item \emph{similarity \reviewtwo{and distance}-based} methods compare whole time series using similarity \reviewtwo{and distance} measures, usually in conjunction with 1-NN classifiers. Particularly, 1-NN with DTW~\cite{sakoe1978dynamic,itakura1975minimum} was long
    considered as the de facto standard for univariate TSC\@. More accurate similarity \reviewtwo{and distance}-based methods combine multiple measures,
    including 1-NN-based ensemble Elastic Ensemble (EE)~\cite{lines2015time},
    and tree-based ensemble Proximity Forest (PF)~\cite{lucas2019proximity}. In Section~\ref{sec:measures} we will
    explore more details of several similarity \reviewtwo{and distance} measures used in TSC.
    \item \emph{Interval-based} methods use summary statistics relating to subseries  in
    conjunction with location information as discriminatory features. Examples include Time Series Forest
    (TSF) \cite{deng2013time},
    Random Interval Spectral Ensemble (RISE) \cite{lines2018time}, Canonical Interval Forest (CIF)
    \cite{middlehurst2020canonical} and Diverse Representation CIF (DrCIF)~\cite{middlehurst2020canonical}. Currently, DrCIF is the most accurate classifier in this category~\cite{middlehurst2021hive}.
    \item \emph{Shapelet-based} methods extract or learn a set of discriminative subseries for each class which are then used as search keys for the particular classes. The presence, absence or distance of a shapelet is used as discriminative information for classification. 
    Examples include Shapelet Transform (ST)
    \cite{hills2014classification} and
    Generalized Random Shapelet Forest (gRSF) \cite{karlsson2016generalized}, and a time contracted version of ST called Shapelet Transform Classifier (STC)~\cite{bagnall2020usage}.
    \item \emph{Dictionary-based} methods transform time series into a bag-of-word model. The series is
    either discretized in time domain such as in Bag of Patterns (BoP)~\cite{lin2012rotation} or it is
    transformed into the frequency domain such as in Bag-of-SFA-Symbols (BOSS) \cite{schafer2015boss}, and
    Word eXtrAction for time SEries cLassification (WEASEL) \cite{schafer2017fast} and Temporal Dictionary Ensemble (TDE)~\cite{middlehurst2020temporal}. Currently, TDE is the most accurate classifier in this category~\cite{middlehurst2021hive}.
    \item \reviewone{\emph{Kernel-based}} methods transform the time series using a
    transformation function and then use a general purpose classifier. A notable example is RandOm
    Convolutional KErnel Transform (ROCKET) \cite{dempster2020rocket} which uses random convolutions to
    transform the data, and then uses logistic
     regression for classification.
    \item \emph{Deep-learning} methods can be divided into two main types of architectures: (1) based on recurrent neural networks \cite{gallicchio2017deep}, or (2) based on temporal convolutions, such as Residual Neural Network
    (ResNet) \cite{wang2017time} and InceptionTime \cite{fawaz2020inceptiontime}. A recent review of deep learning methods shows that architectures that use temporal convolutions show higher accuracy \cite{fawaz2019deep}.
    \item \emph{Combinations of Methods} combine multiple methods to form
    ensembles. Examples include HIVE-COTE (Hierarchical Vote Collective of Transformation-based
    Ensembles)~\cite{lines2018time}, which ensembles EE, ST, RISE and BOSS, and TS-CHIEF
    (Time Series Combination of Heterogeneous and Integrated Embeddings Forest)~\cite{shifaz2020ts}, which
    is a tree-based ensemble where the tree nodes use similarity, distance, dictionary or interval-based splitters.
\end{itemize}


\reviewone{
InceptionTime, TS-CHIEF, ROCKET and HIVE-COTE have been identified to be state-of-the-art classifiers for TSC~\cite{middlehurst2021hive}.
The latest versions of HIVE-COTE, which do not include EE, called HIVE-COTE 1.0~\cite{bagnall2020usage} and HIVE-COTE 2.0~\cite{middlehurst2021hive},  significantly improved the speed of the original version~\cite{lines2018time}. While benchmarking on the UCR univariate TSC archive places HIVE-COTE 1.0 behind ROCKET and TS-CHIEF on accuracy, \cite{bagnall2020usage}, recent benchmarking \cite{middlehurst2021hive} places HIVE-COTE 2.0 ahead on accuracy relative to all alternatives.
}

\subsection{Multivariate TSC}
\label{subsec:multi-tsc}

Research into multivariate TSC has lagged behind univariate research. A recent paper~\cite{ruiz2021great} reviews several methods used for multivariate TSC and compares their performance on the UEA multivariate time series archive~\cite{bagnall2018uea}.
Here, we present a short summary of multivariate methods:

\begin{itemize}
    \item Multivariate \emph{similarity \reviewtwo{and distance} measures} can be used with 1-nearest neighbor for classification. DTW has previously been extended to the multivariate case using two key strategies~\cite{shokoohi2017generalizing}. The \emph{independent} strategy applies the univariate measure to each dimension and then sums the resulting distances. The \emph{dependent} strategy treats each time step as a multi-dimensional point. DTW is then applied on the Euclidean distances between these multidimensional points. Figure~\ref{fig:mdtw}%
\begin{figure*}[ht]
    \centering
    \includegraphics[width=.8\linewidth]{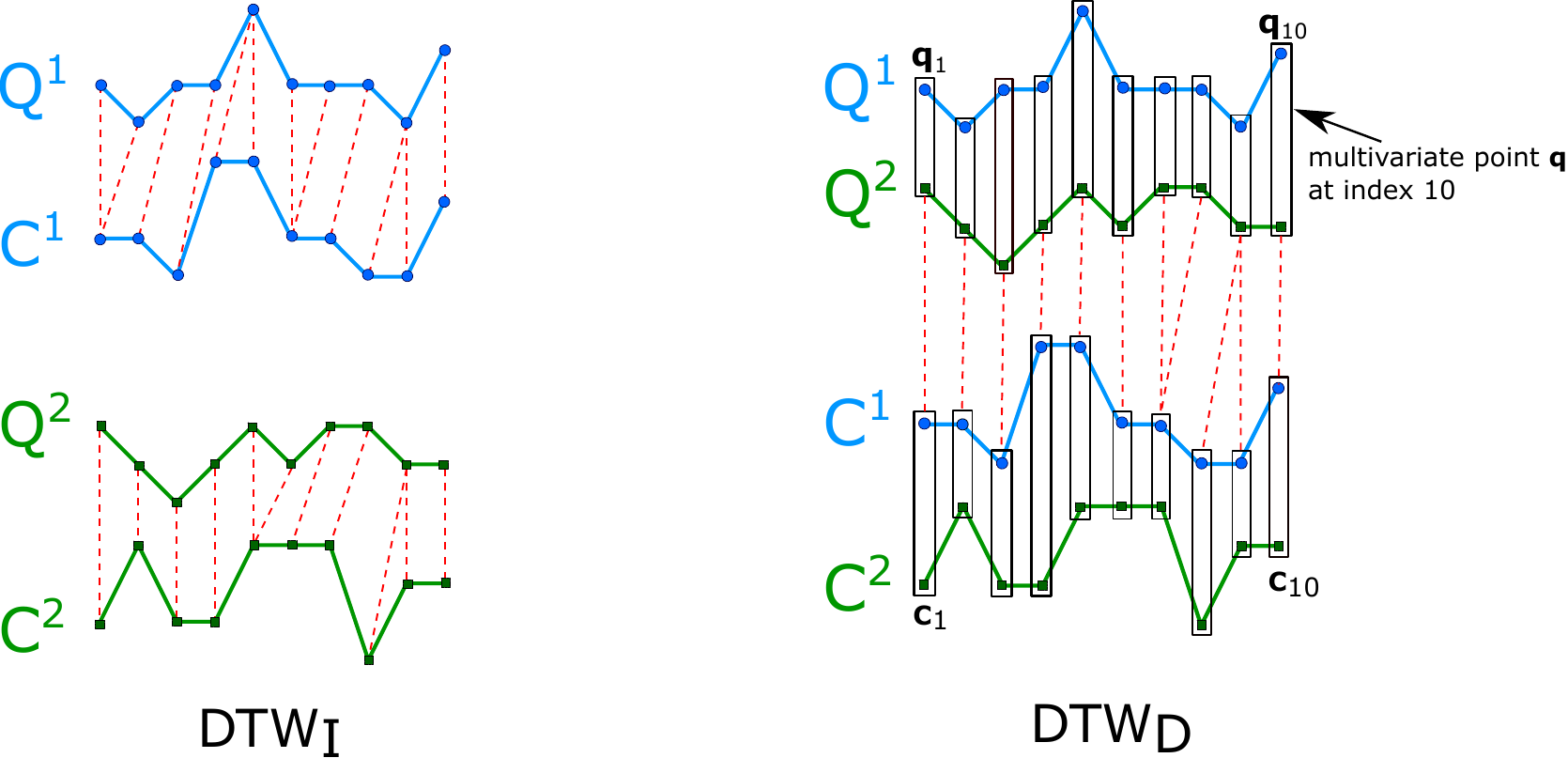}
    \caption{Independent DTW ($DTW_{I}$, left) and dependent DTW ($DTW_{D}$, right). Dimension 1 in series $Q$ and $C$ is shown in blue, and the dimension 2 is shown in green.
    }
    \label{fig:mdtw}
\end{figure*}
    illustrates these approaches and we present definitions in Section~\ref{sec:measures}.
    \reviewone{In addition, Shokoohi et al. \cite{shokoohi2017generalizing} present an adaptive approach to select between independent or dependent DTW based on the performance on the dataset. Since this adaptive approach falls back to either the independent or dependent version based on the performance, we only studied the independent and dependent strategies when experimenting with single measures. However, in Section~\ref{subsec:mee} we present an ensemble based on our own version of an adaptive approach. We discuss its results in Section~\ref{ch6:subsec:measures-vs-mee}.}
    \item \emph{Interval-based} methods include RISE~\cite{lines2018time}, TSF~\cite{deng2013time}, and the recently introduced CIF~\cite{middlehurst2020canonical} and its extension DrCIF~\cite{middlehurst2021hive}. They extract intervals from each dimension separately. CIF and DrCIF have shown promising results for multivariate classification~\cite{ruiz2021great}.%
    \item \emph{Shapelet-based} methods include gRSF~\cite{karlsson2016generalized} and time contracted Shapelet Transform (STC)~\cite{bagnall2020usage}. According to the review~\cite{ruiz2021great}, STC is the current most accurate multivariate method that uses shapelets~\cite{ruiz2021great}. 
    \item \emph{Dictionary-based} methods include WEASEL with
a Multivariate Unsupervised Symbols and dErivatives (MUSE)
(a.k.a WEASEL+MUSE, or simply MUSE)~\cite{schafer2017fast} and
time contracted Bag-of-SFA-Symbols (CBOSS)~\cite{middlehurst2019scalable}.
    \item \emph{Kernel-based} methods include an extension of ROCKET to the multivariate case, implemented in the \emph{sktime} library~\cite{sktime}. It combines information from multiple dimensions using small subsets of dimensions.
    \item \emph{Combinations of Methods} include a multivariate version of HIVE-COTE 1.0 which combines STC, TSF, CBOSS, and RISE~\cite{bagnall2020usage}, applying each constituent algorithm to each dimension separately.
    \item \emph{Deep-learning-based} methods that directly support multivariate series include Time Series Attentional Prototype Network (TapNet)~\cite{zhang2020tapnet}, ResNet~\cite{fawaz2019deep} and InceptionTime~\cite{fawaz2020inceptiontime}. 
\end{itemize}

To benchmark key algorithms covered in the review, Ruiz~et~al.\ compared 12 classifiers on 20 UEA multivariate datasets with equal length that completed in a reasonable time. They found that the most accurate multivariate TSC algorithms are ROCKET, InceptionTime, MUSE, CIF, HIVE-COTE and MrSEQL in that order~\citep[Figure~12a]{ruiz2021great}.


\section{Similarity \reviewtwo{and Distance} Measures}
\label{sec:measures}

In this section we present the proposed similarity \reviewtwo{and distance} measures. For this study, we extend to the multivariate case the
set of univariate similarity \reviewtwo{and distance} measures used in EE and PF (and thus TS-CHIEF and some versions of HIVE-COTE).


The independent strategy proposed by \cite{shokoohi2017generalizing} simply sums over the results of applying DTW separately to each dimension. For completeness we propose to extend this idea to allowing any $p$-norm. In this case, the previous approach extends directly to any univariate measure as follows.
\goodbreak
    \begin{definition}{Independent Measures}

        For any univariate measure $m(Q^1,C^1)\rightarrow \mathbb{R}$ and multivariate series $Q$ and $C$, an independent multivariate extension of $m$ is defined by

        \begin{equation}
            Ind(m, Q, C, p) =
            \left(
            \sum_{d=1}^{D}\left| m(Q^d, C^d)\right|^p
            \right)^{1/p}
            \label{eq:indep}
        \end{equation}
    \end{definition}

   We compute the  distance between $Q$ and $C$ separately for each dimension, and then
    take the $p$-norm of the results.
    Here,
    $Q^d$ (or $C^d$) represents the univariate time series of dimension $d$ such that $Q^d=<q_1^d,\cdots,q_L^d>$ (or $C^d=<c_1^d,\cdots,c_L^d>$).
    The parameter $p$ is set to 1 in Shokoohi-Yekta et al.~\cite{shokoohi2017generalizing}. 

For consistency with previous work, we assume a 1-norm unless otherwise specified. For ease of comprehension, we indicate an independent extension of a univariate measure by adding the subscript $I$. Hence, $DTW_I(Q,C)=Ind(DTW, Q, C, 1)$, $WDTW_I(Q,C)=Ind(WDTW, Q, C, 1)$ and so forth.

However, in most cases it requires more than such a simple formulation to derive a dependent extension, and hence we below introduce each of the univariate measures together with our proposed dependent variant.

\subsection{$Lp$ Distance (Lp)}
\label{subsec:lpd}

\subsubsection{Univariate $Lp$ Distance}
\label{subsubsec:uni-lpd}

The simplest way to calculate distance between two time series is to use $Lp$ distance, also known as the Minkowski distance.

Let us denote by $Q$ and $C$ two univariate ($D = 1$) time series of length $L$ where $q_i$ and $c_i$ are scalar values at time point $i$ from the two time series.
Equation~\ref{eq:uni-minkowski} formulates the $Lp$ distance between $Q$ and $C$.

\begin{equation}
    Lp(Q, C) = \left(\sum_{i}^{L}{\left|q_i - c_i\right|^p}\right)^{1/p}
    \label{eq:uni-minkowski}
\end{equation}

The parameter $p$ is the order of the distance. The $L_1$ 
(Manhattan distance) and $L_2$  (Euclidean distance) distances are widely used.
%
%


\subsubsection{Multivariate $Lp$ Distance}
\label{subsubsec:multi-lpd}

We here show that Independent $Lp$ distance ($Lp_{I}$) and Dependent $Lp$ distance ($Lp_{D}$) are identical for a given value of the parameter $p$.

\begin{definition}{Independent $Lp$ Distance ($Lp_I$)}

    In this case, we simply compute the $Lp$ distance between $Q$ and $C$ separately for each dimension, and then
    take the $p$-norm of the results. 
    

    \begin{equation}
        \begin{split}
            Lp_{I}(Q, C) & =
            \left(
            \sum_{d=1}^D
            \left |
            Lp(Q^d, C^d)
            \right |^{p}
            \right)^{1/p}
            \\& =
            \left(
            \sum_{d=1}^D 
            \sum_{i=1}^{L}\left |q^d_{i} - c^d_{i}\right |^p
            \right)^{1/p}
        \end{split}
        \label{eq:lp-indep}
    \end{equation}

\end{definition}

\begin{definition}{Dependent $Lp$ Distance ($Lp_D$)}

In this case, we compute the $Lp$ distance between each multidimensional point $\mathbf{q_{i}}\in\mathbb{R}^D$ and $\mathbf{c_{i}}\in\mathbb{R}^D$, 
and take the $p$-norm of the results.

    \begin{equation}
        \begin{split}
            Lp_{D}(Q, C) & =
            \left(
            \sum_{i=1}^L
            \left |
            Lp(\mathbf{q_{i}} , \mathbf{c_{i}})
            \right |^{p}
            \right)^{1/p}
            \\ & =
            \left(
            \sum_{i=1}^L 
            \sum_{d=1}^{D}\left|q^d_{i} - c^d_{i}\right |^p 
            \right)^{1/p}
        \end{split}
        \label{eq:lp-dep}
    \end{equation}
\end{definition}

Consequently both the independent and the dependent versions of the \emph{non-elastic} $Lp$ distance will produce the same
result when used with the same value of $p$. 

In the context of TSC, $Lp$ distances are of limited use because they cannot align two series that are misaligned in the time dimension, since they compute one-to-one differences between 
corresponding points only.

For example, in an electrocardiogram (ECG) signal, two measurements from a patient at different times may
produce slightly different time series which belong to the same class \reviewone{(e.g.\ a certain heart condition)}. Ideally,
if they belong to the same class, an effective similarity or distance measure should account for such ``misalignments''
in the time axis, while capturing the similarity or distance.


\textit{Elastic} similarity \reviewtwo{and distance} measures such as DTW tackle this issue. Elastic 
measures are designed to compensate for temporal misalignments in time series
that might be due to
stretched, shrunken or misaligned subsequences. From Section~\ref{subsec:dtw} to~\ref{subsec:twe} we
will present various \textit{elastic} similarity \reviewtwo{and distance} measures, and show that independent and dependent strategies are substantially different.

\subsection{Dynamic Time Warping (DTW) and Related Measures}
\label{subsec:dtw}

\subsubsection{Univariate DTW}
\label{subsubsec:uni-dtw}

The most widely used \textit{elastic} distance measure is DTW \cite{sakoe1978dynamic}. By contrast to measures such as the $Lp$ distance, DTW is an elastic distance measure, that allows one-to-many alignment (``warping'') of points between two time series. 
For many years, 1-NN with DTW was considered as the traditional benchmark
algorithm for TSC \cite{ding2008querying}.

DTW is efficiently solved using a dynamic programming technique. Let $\Delta_{DTW}$ be an $(L+1)$-by-$(L+1)$ dynamic
programming cost matrix with indices starting from $i=0,j=0$. The first row $i=0$ and the first column $j=0$ defines the border conditions:
\begin{equation}
    \begin{matrix}
        \begin{array}{ll}
            \Delta_{DTW}(0,~0) & = 0 \\[2pt]
            \Delta_{DTW}(i,~0) & = +\infty\text{, } 1\leq i\leq L \\[2pt]
            \Delta_{DTW}(0,~j) & = +\infty\text{, } 1\leq j\leq L \\
        \end{array}
    \end{matrix}
    \\
    \label{eq:dtw-matrix-boundary}
\end{equation}

The rest of the elements $(i, j)~\textit{with}~i > 0 ~\textit{and}~j > 0$ is defined as the squared Euclidean distance between two corresponding points $q_i$ and $c_j$ -- \ie $\Delta_{DTW}(i, j) = (q_i - c_j) ^ 2$ and the minimum of the cumulative distances of the previous points. Equation~\ref{eq:dtw-matrix-inner} defines element ($i,j$), where $i>0$ and $j>0$, of the cost matrix as follows:
\begin{equation}    
    \Delta_{DTW}(i,~j) = (q_i - c_j) ^ 2 
     + min
    \left\{
    \begin{matrix}
        \begin{array}{ll}
            \Delta_{DTW}(i-1,j-1)\\
            \Delta_{DTW}(i,j-1) \\
            \Delta_{DTW}(i-1,j)
        \end{array}
    \end{matrix}
    \right.
    \label{eq:dtw-matrix-inner}
\end{equation}

The cost matrix represents the alignment of the two series as according to the DTW algorithm.
DTW between two series $Q$ and $C$ is the accumulated cost in the last element of the cost matrix (\ie{ $i,j = L +1$ }) as defined in Equation~\ref{eq:dtw}:

\begin{equation}
    DTW(Q,C) = \Delta_{DTW}(L,L).
    \label{eq:dtw}
\end{equation}




\reviewone{
Note that, except for LCSS, the boundary conditions of the matrix remain the same for the multivariate measures. Hence we do not repeat the boundary conditions in Equation~\ref{eq:dtw-matrix-boundary} for measures other than LCSS. We simply modify Equation~\ref{eq:dtw-matrix-inner} accordingly to each measure.
}

DTW has a parameter called ``window size'' ($w$), which helps to prevent pathological
warpings by constraining the maximum allowed warping distance.
For example, when $w=0$, DTW produces a one-to-one alignment, which is equivalent to the
Euclidean distance. A larger warping window allows one-to-many alignments where
points from one series can match points from the other series over longer time frames.
Therefore, $w$ controls the \textit{elasticity} of the distance measure.

Different methods have been used to select the parameter $w$.
In some methods, such as EE, and HIVE-COTE, $w$ is selected using leave-one-out cross-validation.
Some algorithms select the window size randomly (\eg PF and TS-CHIEF select window sizes from the uniform distribution $U(0, L/4)$).

Parameter $w$ also improves the computational efficiency, since in most cases, the ideal $w$ is much less
than the length of the series~\cite{tan2018efficient}. When $w$ is small, DTW runs relatively fast, especially with lower bounding,
and early abandoning techniques
~\cite{keogh2009supporting,lemire2009faster,tan2017indexing,herrmann2021early}.
Time complexity to calculate DTW with a warping window is~$O(L \cdot w)$, instead of $O(L^2)$ for the full DTW.

\reviewone{
In this paper, we use DTW to refer to DTW with a cross-validated window parameter and DTWF to refer to DTW with window set to series length.
}

\subsubsection{Dependent Multivariate DTW}
\label{subsubsec:multi-dtw}







\begin{definition}{Dependent DTW ($DTW_D$)}

    Dependent DTW ($DTW_{D}$) uses all dimensions together when computing the
    point-wise distance between each point in the two time series~\cite{shokoohi2017generalizing}.
    In this method, for each point in the series, DTW is allowed to warp across the dimensions.

    In this case, the squared Euclidean distance between two univariate points -- $(q_i - c_j) ^ 2$ -- in
    Equation~\ref{eq:dtw-matrix-inner} is replaced with the $L_2$-norm computed between the two multivariate points $\mathbf{q_i}$ and $\mathbf{c_j}$ as
    in Equation~\ref{eq:dtw-dep}.

    \begin{equation}
        L_2(\mathbf{q_i}, \mathbf{c_j})^2 =  \sum_{d=1}^{D}(q^d_i - c^d_j)^2
        \label{eq:dtw-dep}
    \end{equation}

\end{definition}



\subsubsection{Derivative DTW (DDTW)}
\label{uni-ddtw}
Derivative DTW (DDTW) is a variation of DTW, which computes DTW on the first derivatives of time series. Keogh~et~al. 
\cite{keogh2001derivative} developed this version to mitigate some pathological warpings,
particularly, cases where DTW tries to explain variability in the time series values by warping the time-axis, and cases
where DTW misaligns features in one series which
are higher or lower than its corresponding features in the other series.
The derivative transformation of a univariate time point $q'_{i}$ is defined as:

\begin{equation}
    q'_{i} = \frac{(q_i - q_{i-1} + (q_{i+1} - q_{i-1})/2)}{2}
    \label{eq:ddtw-transform}
\end{equation}

Note that $q'_{i}$ is not defined for the first and last element of the time series. Once the two series have
been transformed, DTW is computed as in Equation~\ref{eq:dtw}.

Multivariate versions of DDTW are very straightforward to implement. We calculate the derivatives separately for each dimension, and then use Equations \ref{eq:indep} and \ref{eq:dtw-dep} to compute from the derivatives independent DDTW ($DDTW_I$) and dependent DDTW ($DDTW_D$), respectively. 

\subsubsection{Weighted DTW (WDTW)}
\label{uni-wdtw}
Weighted DTW (WDTW) is another variation of DTW, proposed by \cite{jeong2011weighted}, which uses a ``soft warping
window'' in contrast to the fixed warping window sized used in classic DTW\@.
WDTW penalises large warpings by assigning a
non-linear multiplicative weight $w$ to the warpings using the modified logistic function in
Equation~\ref{eq:wdtw-weight}:

\begin{equation}
    weight_{|i{-}j|} = \frac{weight_{max}}{1 + e^{-g\cdot((|i{-}j|-L)/2)}},
    \label{eq:wdtw-weight}
\end{equation}
where $weight_{max}$ is the upper bound on the weight (set to 1), $L$ is the series length and $g$ is the
parameter that
controls the penalty level for large warpings.
Larger values of $g$ increases the penalty for warping.

When creating the dynamic programming distance matrix $\Delta_{WDTW}$, the weight penalty $weight_{|i{-}j|}$
for a warping
distance of $|i{-}j|$ is applied, so that the $(i,j)$-th ground cost in the matrix $\Delta_{\mathit{WDTW}}$ is
$weight_{|i{-}j|}~\cdot~(q_i~-~c_i)^2$.
Therefore, the new equation for WDTW is defined as

\begin{multline}
    \Delta_{\mathit{WDTW}}(i, j) = weight_{|i{-}j|} \cdot  (q_i - c_j) ^ 2   \\
    +min
    \left\{
    \begin{matrix}
        \begin{array}{ll}
            \Delta_{\mathit{WDTW}}(i-1,j-1)\\
            \Delta_{\mathit{WDTW}}(i,j-1)\\
            \Delta_{\mathit{WDTW}}(i-1,j)
        \end{array}
    \end{matrix}
    \right.
    \label{eq:wdtw-matrix}
\end{multline}

\begin{equation}
    \mathit{WDTW}(Q,C) = \Delta_{\mathit{WDTW}}(L, L).
    \label{eq:wdtw}
\end{equation}

Parameter $g$ may be selected using leave-one-out cross-validation as in EE and HIVE-COTE, or 
selected randomly as in PF and TS-CHIEF ($g\sim U(0,1)$).

Since WDTW does not use a constrained warping window (\ie the maximum warping distance $|i{-}j|$ may be as
large as $L$), its time complexity is $O(L^2)$, which is higher than DTW\@.

\subsubsection{Dependent Multivariate WDTW}
\label{subsubsec:multi-wdtw}





\begin{definition}{Dependent WDTW}

    The dependent version of WDTW simply inserts the weight into $DTW_D$. We define Dependent WDTW ($\mathit{WDTW}_{D}$) as,

    \begin{multline}
        \Delta_{\mathit{WDTW}_{D}}(i, j) = weight_{|i{-}j|} \cdot L_2(\mathbf{q_i}, \mathbf{c_{j}})^2 
        \\ \quad + \; min \left\{
        \begin{matrix}
            \begin{array}{ll}
                \Delta_{\mathit{WDTW}_{D}}(i-1,j-1)\\
                \Delta_{\mathit{WDTW}_{D}}(i-1,j)\\
                \Delta_{\mathit{WDTW}_{D}}(i,j-1),
            \end{array}
        \end{matrix}
        \right.
        \label{eq:wdtw-dep-matrix}
    \end{multline}

\begin{equation}
    \mathit{WDTW}_D(Q,C) = \Delta_{\mathit{WDTW}_{D}}(L,L).
    \label{eq:wdtw-dep}
\end{equation}
\end{definition}

\subsubsection{Weighted Derivative DTW (WDDTW)}
\label{uni-wddtw}

The ideas behind DDTW and WDTW may be combined to implement another measure called Weighted Derivative DTW
(WDDTW). This method has also been traditionally used in some ensemble algorithms~\cite{bagnall2017great}.

Multivariate versions of WDDTW are also straightforward to implement. We calculate the derivatives separately for each dimension, and then use Equations \ref{eq:indep} and \ref{eq:wdtw-dep-matrix} with them to compute independent WDDTW ($\mathit{WDTW}_I$) and dependent WDDTW ($\mathit{WDTW}_D$), respectively. 

\subsection{Longest Common Subsequence (LCSS)}
\label{subsec:lcss}

\subsubsection{Univariate LCSS}
\label{subsubsec:uni-lcss}

\begin{figure}[h]
    \centering
    \includegraphics[width=.6\linewidth]{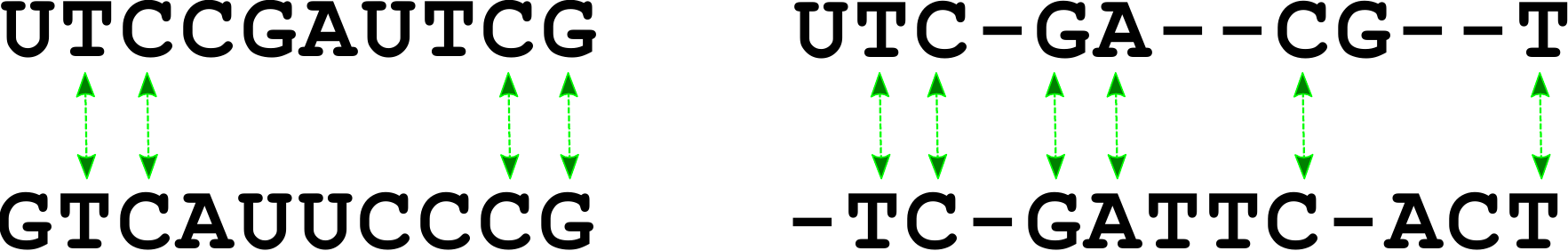}
    \caption{Example of string matching with LCSS. Image on the left side shows direct pairwise matching of
     two strings using LCSS. Image on the right side shows matching of two strings with some gaps or unmacthed letters (shown as ``--'') allowed between the matched letters. This allows matching with ``elascticity'' as in DTW.}
    \label{fig:lcss-example}
\end{figure}

Longest Common Subsequence (LCSS) distance is based on the edit distance algorithm, which is used for string matching \cite{hirschberg1977algorithms,vlachos2002discovering}.
Figure~\ref{fig:lcss-example} shows an example string matching problem. 
\reviewone{
One of the early works that use LCSS for time series classification by Vlachos et al.~\cite{vlachos2002discovering} states that one motivation to use LCSS based approach is that it is more robust to noise compared to DTW. This is because DTW matches all points, including the outliers. However, LCSS can allow some points to remain unmatched while retaining the order of matching. In addition LCSS is designed to be more efficient than DTW since it does not require the $Lp$ computation.
}
In TSC, the LCSS algorithm is
modified to work with real-valued data by adding a threshold $\epsilon$ for real-value comparisons.
\reviewone{Two real-values are considered a match if the difference between them is not larger than the threshold $\epsilon$.
}
A warping window can also be used in conjunction with the threshold to constrain the degree of local warping.

The unnormalized LCSS distance ($LCSS_{UN}$) between $Q$ and $C$ is




\begin{multline}
\\
    \Delta_{LCSS}(0,~0) = 0 \text{ and } \Delta_{LCSS}(i,~0) = 
    \Delta_{LCSS}(0,~j) = -\infty\text{, } 1\leq i,j \leq L \\
    \Delta_{LCSS}(i, j) =
    \left\{
    \begin{matrix}
        \begin{array}{ll}
            1 + \Delta_{LCSS} (i-1, j-1) & \textbf{if } |q_i - c_j| \leq \epsilon \\
            max \left\{
            \begin{matrix}
                \Delta_{LCSS} (i-1, j) \\
                \Delta_{LCSS} (i, j-1)
            \end{matrix}\right. & \textbf{otherwise},\\
        \end{array}
    \end{matrix}\right.
    \label{eq:lcss-matrix}
\end{multline}

\begin{equation}
    LCSS_{UN}(Q,C) = \Delta_{LCSS} (L,L),
    \label{eq:lcss-unnorm}
\end{equation}

In practice, $LCSS_{UN}$ is then normalized based on the series length $L$.

\begin{equation}
    LCSS(Q,C) = 1 - \frac{LCSS_{UN} (Q,C)} {L},
    \label{eq:lcss}
\end{equation}

LCSS can be used with a window parameter $w$ similar to DTW\@.
With a window parameter, LCSS has a time complexity of~$O(L \cdot w)$.
In EE and PF, the parameter $\epsilon$ is selected from $[\frac{\sigma}{5},\sigma]$, where $\sigma$ is the standard
deviation of the whole dataset. 

\subsubsection{Independent Multivariate LCSS}
\label{subsubsec:indep-lcss}

Independent multivariate $\mathit{LCSS_{I}}$ uses the Equation~\ref{eq:indep} and computes the LCSS for dimensions separately. However, in this case we use a separate $\epsilon$ parameter for each dimension. We compute the standard deviation $\sigma$ per dimension when sampling $\epsilon$. Sampling $\epsilon$ for each dimension can be useful if the data is not normalized.

\subsubsection{Dependent Multivariate LCSS}
\label{subsubsec:dep-lcss}





\begin{definition}{Dependent LCSS}

    \reviewone{Dependent LCSS ($LCSS_{D}$) is similar to Equation~\ref{eq:lcss-matrix}, except that to compute
    distance between two multivariate points
    we use Equation~\ref{eq:dtw-dep} and an adjustment is made to the range of parameter $\epsilon$ to make allowance for multidimensional distances tending to be larger than univariate.
     In this case, parameter $\epsilon$ is selected from $[\frac{2\cdot D\cdot \sigma}{5},2\cdot D\cdot \sigma]$, where $\sigma$ the standard deviation of the whole dataset. Except for this adjustment, parameters are sampled similar to the way it was selected in the univariate LCSS, in EE.}

    \begin{multline}
        \Delta_{LCSS_{D}}(i, j) =
        \left\{
        \begin{matrix}
            \begin{array}{ll}
                1 + \Delta_{LCSS_{D}} (i{-}1, j{-}1) & \textbf{if } L_2(\mathbf{q_i},
                \mathbf{c_j})^2 \leq \epsilon \\
                max \left\{
                \begin{matrix}
                    \Delta_{LCSS_{D}} (i{-}1, j) \\
                    \Delta_{LCSS_{D}} (i, j{-}1)
                \end{matrix}\right. & \textbf{otherwise},\\
            \end{array}
        \end{matrix}\right.
        \label{eq:lcss-dep-matrix}
    \end{multline}
    
    
    \begin{equation}
        LCSS_{UND}(Q,C) = \Delta_{LCSS_{D}} (L, L),
        \label{eq:lcss-dep-unnorm}
    \end{equation}
    
    Similar to the univariate case, $LCSS_{UND}$ is then normalized based on the series length $L$.
    
    \begin{equation}
        LCSS_{D}(Q,C) = 1 - \frac{LCSS_{UND} (Q,C)} {L}.
        \label{eq:lcss-dep}
    \end{equation}

\end{definition}

    \label{subsub:param-lcss}

\reviewone{
\subsubsection{Other $LCSS_{D}$ Formulations}
\label{vlachos-lcss-dependent}

An early work by Vlachos~et~al.~\citep{vlachos2003indexing} also presented a way to extend measures to the multivariate case. 
Their proposed dependent DTW is similar to Shokoohi~et~al.'s ${DTW_D}$ formulation in Equation~\ref{eq:dtw-dep}, but their LCSS's formulation is slightly different to our LCSS formulation present in Equation~\ref{eq:lcss-dep-matrix}. Equation~\ref{eq:vlachos-lcss-dep-matrix} defines this version of LCSS named $Vlachos\_LCSS_{D}$. 

    \begin{multline}
        \Delta_{Vlachos\_LCSS_{D}}(i, j) =
        \left\{
        \begin{matrix}
            \begin{array}{ll}
                1 + \Delta_{LCSS_{D}} (i{-}1, j{-}1) & \textbf{if } \forall~d~\in~D, |q^d_i - c^d_j| < \epsilon \\ &
                \textbf{and } |i - j| \leq \delta \\
                max \left\{
                \begin{matrix}
                    \Delta_{LCSS_{D}} (i{-}1, j) \\
                    \Delta_{LCSS_{D}} (i, j{-}1)
                \end{matrix}\right. & \textbf{otherwise},\\
            \end{array}
        \end{matrix}\right.
        \label{eq:vlachos-lcss-dep-matrix}
    \end{multline}

Our method treats $\mathbf{q_i}$ and $\mathbf{c_i}$ as each being multi-dimensional points, placing a single constraint on the distance between them, whereas Vlachos~et~al. treat the dimensions independently, with separate constraints on each. We believe that our method is more consistent with the spirit of dependent measures.

$Vlachos\_LCSS_{D}$ allows matching values within both data dimensions and the time dimension, while our $LCSS_{D}$ matches values only in data dimensions. They have an additional parameter $\delta$ that controls matching in the time dimension.

Our method also requires an adjustment to LCSS's $\epsilon$ parameter as described in Section~\ref{subsub:param-lcss}. This may help to cater for unnormalized data, since $Vlachos\_LCSS_{D}$ use the one $\epsilon$ across all dimensions.



}

\subsection{Edit Distance with Real Penalty (ERP)}
\label{subsec:erp}

\subsubsection{Univariate ERP}
\label{subsubsec:uni-erp}

    Edit Distance with Real Penalty (ERP)~\cite{chen2004marriage,chen2005robust} is also based on string matching
    algorithms.
    In a typical string matching algorithm, two strings, possibly of different lengths, may be aligned
    by doing the least number of add, delete or change operations on the symbols.
    When aligning two series of symbols, the authors proposed that the delete operations in one series can
     be thought of as adding a special symbol to the other series. Chen~et~al.~\cite{chen2004marriage} refers to these added symbols as a ``gap'' element.

    ERP uses the Euclidean distance between elements when there is no gap, and a constant penalty when there
    is a gap.
    This penalty parameter for a gap is denoted as $g$ (see Equation~\ref{eq:erp-matrix}).

    For time series, with real values, similar to the parameter $\epsilon$ in LCSS, a floating point comparison
    threshold
    may be used to determine a match between two values.
    This idea was used in a measure called Edit Distance on Real sequences (EDR)~\cite{chen2004marriage}.
    However, using a threshold breaks the triangle inequality.
    Therefore,  the same authors proposed a variant, namely ERP, which is a measure that follows the triangle
    inequality.
    \reviewone{Being a metric gives some advantages to ERP over DTW or LCSS in tasks such as indexing and clustering.}

    ERP can also be used with a window parameter $w$ similar to DTW\@.
    With the window parameter, ERP has the same time complexity as DTW\@.
    The parameter $g$ is selected from $[\frac{\sigma}{5},\sigma]$, with $\sigma$ being the standard
    deviation of the training data.
    Formally, ERP is defined as,

    \begin{equation}
        \Delta_{ERP}(i, j) = min \left\{
        \begin{matrix}
            \begin{array}{ll}
                \Delta_{ERP}(i-1,j-1) + (q_i - c_j)^2 \\
                \Delta_{ERP}(i-1,j) + (q_i - g)^2 \\
                \Delta_{ERP}(i,j-1) + (c_j - g)^2
            \end{array}
        \end{matrix}
        \right.
        \label{eq:erp-matrix}
    \end{equation}
    
    \begin{equation}
    ERP(Q,C) = \Delta_{ERP} (L, L).
    \label{eq:erp}
    \end{equation}

\subsubsection{Dependent ERP}
\label{subsubsec:multi-erp}




    \begin{definition}{Dependent ERP}

        We define Dependent ERP ($ERP_{D}$) as,

        \begin{equation}
            \Delta_{ERP_{D}}(i, j) = min \left\{
            \begin{matrix}
                \begin{array}{ll}
                    \Delta_{ERP_{D}}(i-1,j-1) + L_2(\mathbf{q_i}, \mathbf{c_{j}})^2 \\
                    \Delta_{ERP_{D}}(i-1,j) + L_2(\mathbf{q_i}, \mathbf{g})^2 \\
                    \Delta_{ERP_{D}}(i,j-1) + L_2(\mathbf{c_j}, \mathbf{g})^2,
                \end{array}
            \end{matrix}
            \right.
            \label{eq:erp-dep-matrix}
        \end{equation}

    \begin{equation}
        ERP_{D}(Q,C) = \Delta_{ERP_{D}}(L, L).
        \label{eq:erp-dep}
    \end{equation}

    In Equation~\ref{eq:erp-dep-matrix}, we note that the parameter $\mathbf{g}$ is a vector that is sampled separately for each dimension.
    This is in contrast to the univariate case in Equation~\ref{eq:erp-matrix} which uses the standard deviation of the whole training dataset (parameter $g$).

    In this case, all terms increase proportionally with respect to the increase in the number of dimensions. So we do not need to adjust for the
    parameter $\mathbf{g}$ as
    we adjusted for $\epsilon$ in LCSS in Section~\ref{subsubsec:dep-lcss}.

    \end{definition}

\subsection{Move-Split-Merge (MSM)}
\label{subsec:msm}

\subsubsection{Univariate MSM}
\label{subsubsec:uni-msm}

    Move-Split-Merge (MSM) is \reviewone{an edit distance-based distance measure} introduced by~\cite{stefan2012move}. The motivation is to
    propose a distance measure that is a metric invariant to translations and robust to temporal misalignments.
    Measures such as DTW and LCSS are not metrics because they fail to satisfy the triangle inequality. 
    \reviewone{Invariance to translation is another design feature of MSM when compared to ERP (see Section~\ref{subsubsec:uni-erp})~\cite{stefan2012move}. For example consider two series $X$ and $Y$ where $X = <v_0 \cdots v_{100}>$ and $Y = <v>$, with $v$ being a constant real value. ERP requires 99 deletes to transform $X$ into $Y$, and the cost of deletion is tied to the value of $v$, since the cost of deletion is 0 if $v=0$ or 99$v$ otherwise (\ie 99 merge). By contrast, in MSM, the cost of deletion is independent of the value of $v$. This makes MSM translation-invariant since it does not change when the same constant is added to two time series~\cite{stefan2012move}.}

\begin{figure}[h]
    \centering
    \includegraphics[width=.6\linewidth]{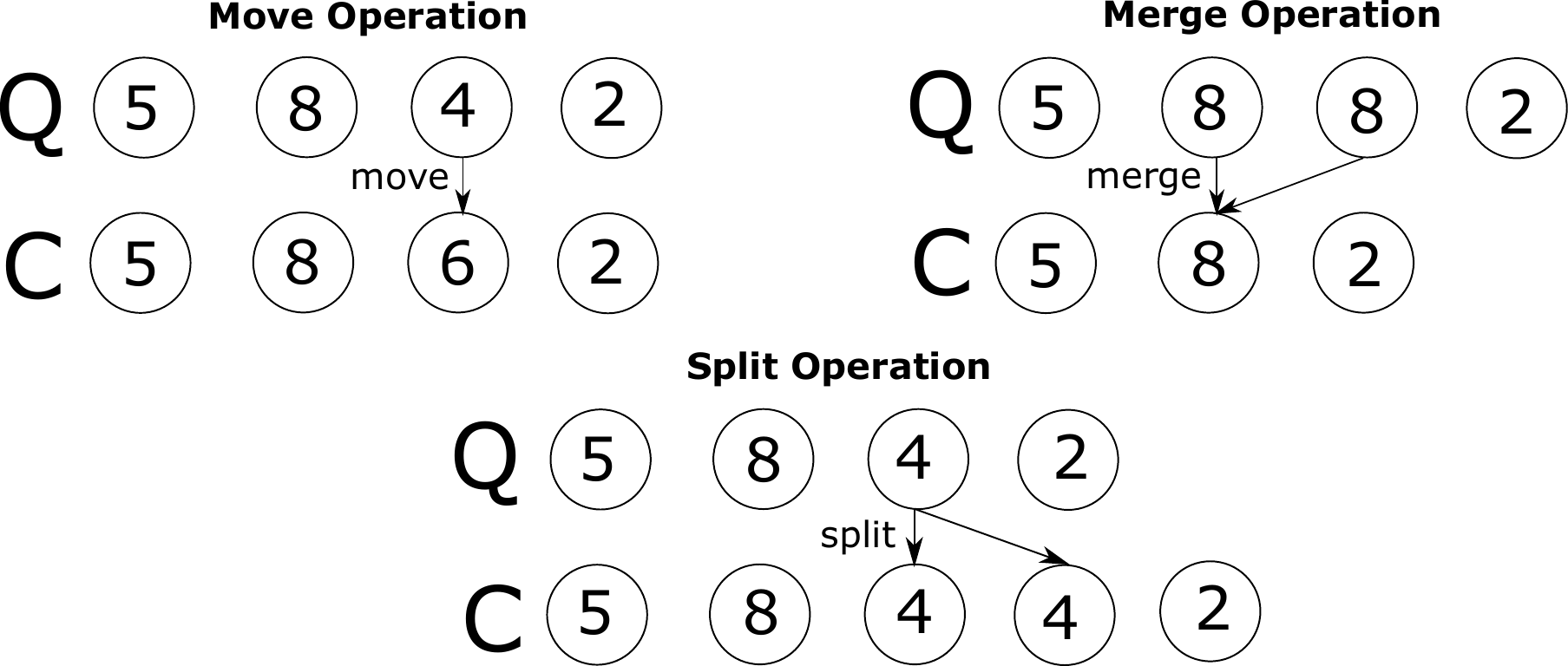}
    \caption{Three edit operations used by MSM to transform one series to the other. MSM attempts to use the minimum number of these operations to perform the transformation. The number of operations used quantifies the distance between two series. Both split and merge operations incur a cost defined by the parameter $c$. Inspired by~\cite{stefan2012move}.}
    \label{fig:msm-ops}
    \end{figure}

    The distance between two series is computed based on
    the number
    and type of edit operations required to transform one series to the other. 
    MSM defines three types of
    edit
    operations: move, merge and split. The move operation substitutes one value into another value. The split operation inserts a copy of the value immediately after itself, and the merge operation is used to delete a value if it directly follows an identical value. Figure~\ref{fig:msm-ops} illustrates these edit operations.
    
    The cost for a
    move operation is
    the pairwise distance between two points, and the
    cost of split or merge operation depends on the parameter $c$.

    Formally, MSM is defined as,

    \begin{equation}
       \Delta_{MSM} (i, j) = min \left\{
        \begin{matrix}
            \begin{array}{ll}
                \Delta_{MSM}(i{-}1,j{-}1) + |q_i - c_j| \\
                \Delta_{MSM}(i{-}1,j) + \textit{cost}(q_i, q_{i-1}, c_j, c) \\
                \Delta_{MSM}(i,j{-}1) + \textit{cost}(c_j, q_i, c_{i-1},c),
            \end{array}
        \end{matrix}
        \right.
        \label{eq:msm-matrix}
    \end{equation}
    
    \begin{equation}
    MSM(Q,C) = \Delta_{MSM} (L , L).
    \label{eq:msm}
    \end{equation}
    The costs of split and merge operations are defined by Equation~\ref{eq:msm-cost}. In the univariate case, the algorithm either merges two values or splits a value if the the value of a point $q_i$ \textit{is\_between} two adjacent values ($q_{i-1}$ and $c_j$). 

    \begin{multline}
        \textit{cost}(q_{i-1},q_i,c_j,c) =
        \left\{
        \begin{matrix}
            \begin{array}{ll}
                c \textbf{ if } q_{i-1} \leq q_i \leq c_j \\
                c \textbf{ if } q_{i-1} \geq q_i \geq c_j \\
                c + min \left\{
                \begin{matrix}
                    \begin{array}{ll}
                        |q_i - q_{i-1}| \\
                        |q_i - c_j|
                    \end{array}
                \end{matrix}
                \right. \textbf{ otherwise}.
            \end{array}
        \end{matrix}
        \right.
        \label{eq:msm-cost}
    \end{multline}

\reviewone{
    In most algorithms (\eg EE, PF, HIVE-COTE and TS-CHIEF), the cost parameter $c$ for MSM is sampled from 100 values generated from 
     the exponential
    sequence
    $\{10^{-2}, \cdots , 10^{2}\}$ proposed in Stefan et al.~\cite{stefan2012move}.
}

\subsubsection{Dependent Multivariate MSM}
\label{subsubsec:multi-msm}





    \begin{definition}{Dependent MSM}

        Here we combine Equation~\ref{eq:msm-matrix} and Equation~\ref{eq:dtw-dep}. The $\textit{cost\_multiv}$ function
        is explained in Section~\ref{subsub:msm-cost}, and presented in Algorithm~\ref{alg:msm-cost-multiv}.

        \begin{multline}
           \Delta_{MSM_{D}}(i,j) = 
            min \left\{
            \begin{matrix}
                \begin{array}{ll}
                    \Delta_{MSM_{D}}(i-1,j-1) + L_2(\mathbf{q_i}, \mathbf{c_j})^2 \\
                    \Delta_{MSM_{D}}(i-1,j) + \mathit{cost\_multiv }(\mathbf{q_i}, \mathbf{q_{i-1}}, \mathbf{c_j}, c) \\
                    \Delta_{MSM_{D}}(i,j-1) + \mathit{cost\_multiv }(\mathbf{c_j}, \mathbf{q_i}, \mathbf{c_{j-1}}, c)
                \end{array}
            \end{matrix}
            \right.
            \label{eq:msm-dep-matrix}
        \end{multline}
        
    \begin{equation}
        MSM_{D}(Q,C) = \Delta_{MSM_{D}}(L, L)
        \label{eq:msm-dep}
    \end{equation}

        \subsubsection{Cost function for dependent MSM}
        \label{subsub:msm-cost}
        
        A nontrivial issue when deriving a dependent variant of MSM is how to translate the concept of one point being between two others.

        A naive approach would test whether a point $x$ \textit{is\_between} points $y$ and $z$ in multidimensional space by
        projecting
        $x$ onto the hyperplane defined by $y$ and $z$.
        However, this has serious limitations.
        For an intuitive example, let us use cities to represent points on a 2-D plane.
        Assume that we have two query cities Chicago and Santiago  wish to determine which
        is between New York and San Francisco.
        If we use vector projections, and project the position of Santiago on to the line between New York and
        San Francisco
        we will find that it is between them.
        Similarly, we will also find that Chicago is between New York and
        San Francisco using this method.
        However, orthogonally Santiago is extremely far away from both New York and
        San Francisco, so it would seem
        more intuitive to define
        this function in a way that Chicago is in between New York and
        San Francisco, but Santiago is not.
        Using this intuition we define the cost function in such a way that a point is considered to be in
        between two points only if
        the point is ``inside the hypersphere'' defined by the other two points.
        Figure~\ref{fig:msm-example} illustrates this concept using three points. 

        
        We implement this idea in Algorithm~\ref{alg:msm-cost-multiv}. First we find the diameter of the hypersphere in line 1 by computing $||\mathbf{q_{i-1}} - \mathbf{c_j}||$. In line 2 we find the mid point $\mathbf{mid}$ along the line $\mathbf{q_{i-1}}$ and $\mathbf{c_j}$. Then we calculate distance to the mid point using $||\mathbf{mid} - \mathbf{q_i}||$ (line 3). Once we have the $distance\_to\_mid$, we check if this distance is larger than half the diameter. If its larger, then the point $\mathbf{q_i}$ is outside the hypersphere, and so we return $c$ (line 5). If $distance\_to\_mid$ is less than half the diameter, then $\mathbf{q_i}$ is inside the hypersphere, so we check to which point
        \reviewone{(either $\mathbf{q_{i-1}}$ or $\mathbf{c_j}$) is closest.} Then we return $c$ plus the distance to the closest point as the cost of the edit operation (line 9 to 12).

        \begin{figure}
            \centering
            \includegraphics[width=.6\linewidth]{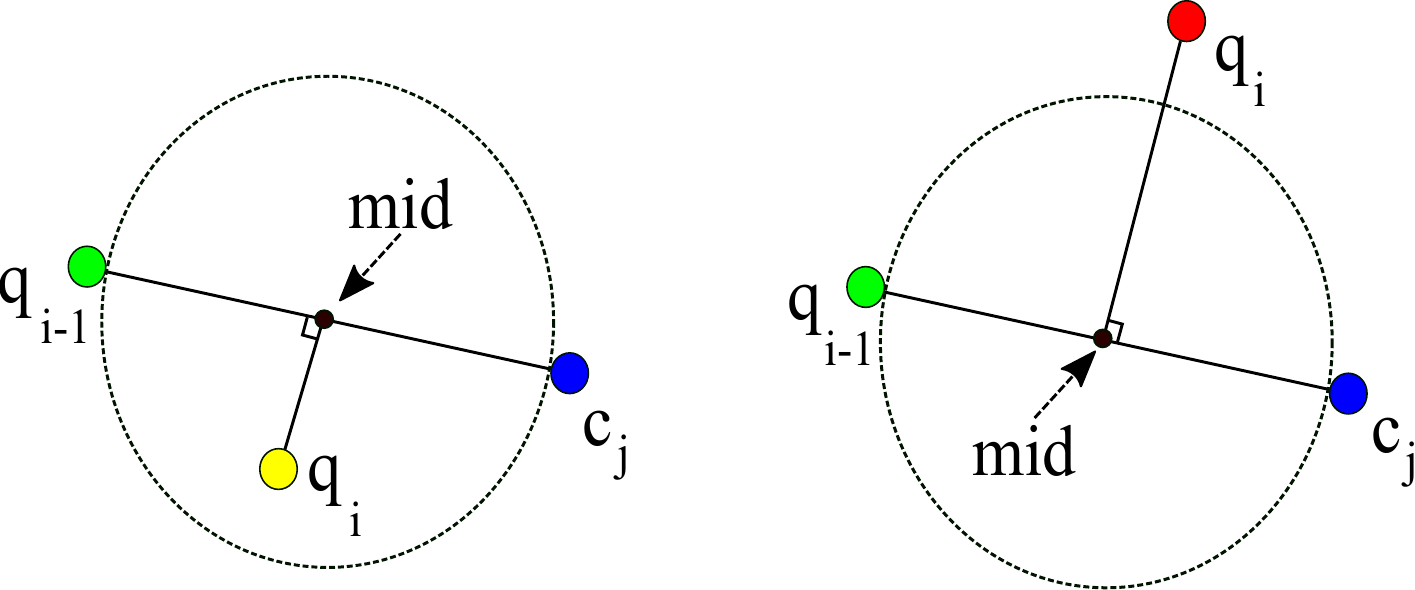}
            \caption{
            Example of checking whether a point is between two other points in 2 dimensions using a
            circle. In the first case (left side), the yellow point is considered ``in between'' blue and
            green points. In the second case (right side), the red point is considered to be ``not in
            between'' the green and blue points even though its projection falls between red and blue points because
            orthogonally it is outside the circle defined by theses points.
            This is one way to
            adapt the
            idea of checking if a point is between
            two other points in the univariate case (1-dimension) as defined in Equation~\ref{eq:msm-cost}.
            In Algorithm~\ref{alg:msm-cost-multiv} we use a generalization of this idea and check
            if a point is inside a hypersphere in D-dimensions.
            }
            \label{fig:msm-example}
        \end{figure}

        \begin{algorithm}
            \label{alg:msm-cost-multiv}
            \SetAlgoLined
            \KwIn{\textit{cost\_multiv}($\mathbf{q_i}, \mathbf{q_{i-1}}, \mathbf{c_j}, c$)
            : three points, and  cost parameter $c$ for MSM}
            \KwOut{cost of operation}
            $diameter = ||\mathbf{q_{i-1}} - \mathbf{c_j}||$\;

            $\mathbf{mid} = (\mathbf{q_{i-1}} + \mathbf{c_j}) / 2$\;
            $distance\_to\_mid = ||\mathbf{mid} - \mathbf{q_i}||$\;

            \uIf{$distance\_to\_mid \leq (diameter / 2)$}{
            \Return $c$\;
            }
            \uElse{
            $dist\_to\_q\_prev = || \mathbf{q_{i-1}} - \mathbf{q_i} ||$\;
            $dist\_to\_c = || \mathbf{c_j} - \mathbf{q_i} ||$\;
            \uIf{$dist\_to\_q\_prev < dist\_to\_c$}{
            \Return $c + dist\_to\_q\_prev$\;
            }
            \uElse{
            \Return $c+ dist\_to\_c$ \;
            }
            }
            \caption{Cost of checking if a mid point is inside the hypersphere defined by the other two points}
        \end{algorithm}

    \end{definition}

\subsection{Time Warp Edit (TWE)}
\label{subsec:twe}

\subsubsection{Univariate TWE}
\label{subsubsec:uni-twe}

    Time Warp Edit (TWE)~\cite{marteau2008time} is a further edit-distance based algorithm adapted to the time series
    domain.
    The goal is to combine an $Lp$ distance based technique with an edit-distance based algorithm that
    supports warping in the time axis, \ie has some sort of \textit{elasticity} like DTW, while
    also being a distance metric (\ie it respects the triangle inequality).
    Being a metric helps in time series indexing, since it speeds up time series retrieval process. 

    TWE uses three operations named \textit{match}, \textit{$delete_A$}, and \textit{$delete_B$}.
    If there is a \textit{match}, $Lp$ distance is used, and if not, a constant penalty $\lambda$ is added.
    \textit{$delete_A$} (or \textit{$delete_B$}) is used to remove an element from the first (or second) series to match the second (or first) series.
    Equations~\ref{eq:twe-matrix},~\ref{eq:twe} and~\ref{eq:twe-ops} define TWE and these three operations, respectively.

    \begin{equation}
        \Delta_{TWE}(i,j) = min \left\{
        \begin{matrix}
            \begin{array}{ll}
                \Delta_{TWE}(i-1,j-1) + \gamma_M & match\\
                \Delta_{TWE}(i-1,j) + \gamma_A & delete_A\\
                \Delta_{TWE}(i,j-1) + \gamma_B & delete_B
            \end{array}
        \end{matrix}
        \right.
        \label{eq:twe-matrix}
    \end{equation}
    
    \begin{equation}
    TWE(Q,C) = \Delta_{TWE} (L, L)
    \label{eq:twe}
    \end{equation}

    

    \begin{equation}
        \begin{matrix}
            \begin{array}{llr}
                \gamma_M = (q_i- c_{j})^2 + (q_{i-1}- c_{j-1})^2
                + 2 \cdot \nu & \;  \  match\\
                \gamma_A = (q_i- q_{i-1})^2 + \nu + \lambda & \;   \ delete_A\\
                \gamma_B = (c_j- c_{j-1})^2 + \nu + \lambda & \;   \ delete_B\\
            \end{array}
        \end{matrix}
        \label{eq:twe-ops}
    \end{equation}


    The multiplicative penalty $\nu$\footnote{In the published definition of TWE~\cite{marteau2008time}, $\nu$ is multiplied with the time difference in the timestamps of two consecutive time points. We simplified this equation, for clarity, by assuming that this time difference is always 1 (UEA datasets do not contain the actual timestamps).} is called the \textit{stiffness} parameter.
    When $\nu = 0$, TWE becomes more stiff like the $Lp$ distance, and
    when $\nu = \infty$, TWE becomes less stiff and more elastic like DTW. The second parameter $\lambda$ is the cost of performing either a \textit{$delete_A$} or \textit{$delete_B$} operation.

\reviewone{
    Following~\cite{marteau2008time, lines2015time, lucas2019proximity}, $\lambda$ is selected from
    $\cup_{i=0}^{9} \frac{i}{9}$ and $\nu$ from the exponentially growing
    sequence $\{10^{-5},5\cdot 10^{-5}, 10^{-4},5 \cdot 10^{-4},10^{-3},5 \cdot 10^{-3},\cdots,1\}$, resulting in 100
    possible parameterizations.
    }
    
\subsubsection{Dependent TWE}
\label{subsubsec:multi-twe}




    \begin{definition}{Dependent TWE}
    
            Dependent version of TWE follows a similar pattern. \reviewone{Due to the greater magnitude of multidimensional distances, $\lambda$ is selected from
    $\cup_{i=0}^{9} \frac{2\cdot D\cdot i}{9}$ and $\nu$ from the exponentially growing
    sequence $\{2\cdot D\cdot 10^{-5},D\cdot 10^{-4},2\cdot D\cdot  10^{-4},D\cdot  10^{-3},2\cdot D\cdot 10^{-3},D\cdot  10^{-2},\cdots,2\cdot D\}$} 
    
    We define Dependent TWE ($TWE_{D}$) as,
            
        \begin{multline}
            \Delta_{TWE_{D}}(i,j) = 
             min \left\{
            \begin{matrix}
                \begin{array}{ll}
                    \Delta_{TWE_{D}}(i-1,j-1) + \gamma_M & \;  \ match\\
                    \Delta_{TWE_{D}}(i-1,j) + \gamma_A & \;  \ delete_A\\
                    \Delta_{TWE_{D}}(i,j-1) + \gamma_B & \;  \ delete_B
                \end{array}
            \end{matrix}
            \right.
            \label{eq:twe-dep-matrix}
        \end{multline}
        
    \begin{equation}
        TWE_{D}(Q,C) = \Delta_{TWE_{D}}(L, L).
        \label{eq:twe-dep}
    \end{equation}


        \begin{multline}
            \begin{matrix}
                \begin{array}{lll}
                    \gamma_M = & L_2(\mathbf{q_i}, \mathbf{c_j})^2 + L_2(\mathbf{q_{i-1}}, \mathbf{c_{j-1}})^2
                     + \; (2 \cdot \nu)   & \quad match\\
                    \gamma_A = & L_2(\mathbf{q_{i}}, \mathbf{q_{i-1}})^2
                    + (\nu + \lambda)  & \quad delete_A\\
                    \gamma_B = & L_2(\mathbf{c_{j}}, \mathbf{c_{j-1}})^2
                    + (\nu + \lambda)  & \quad delete_B\\
                \end{array}
            \end{matrix}
            \label{eq:twe-dep-ops}
        \end{multline}

    \end{definition}


\reviewone{

\subsection{Multivariate Elastic Ensemble (MEE)}
\label{subsec:mee}

Ensembles formed using multiple 1-NN classifiers with a diversity of similarity \reviewtwo{and distance} measures have proved to be significantly more accurate than 1-NN with any single  measure~\cite{lines2015time}. Such ensembles help to reduce the variance of the model and thus help to improve the overall classification accuracy. For example, Elastic Ensemble (EE) combines eleven 1-NN algorithms, each using one of the eleven elastic measures \citep{lines2015time}. 
The eleven measures used in EE are: Euclidean, DTWF (with full window), DTW (with leave-one-out cross-validated window), DDTWF, DDTW, WDTW, WDDTW, LCSS, ERP, MSM, TWE. For each measure, the parameters are optimized with respect to accuracy using leave-one-out cross-validation \citep{lines2015time, bagnall2017great}. Although EE is a relatively accurate classifier \citep{bagnall2017great}, it is slow to train due to the high computational cost of the leave-one-out cross-validation used to tune its parameters -- $O(n^2 \cdot L^2 \cdot P)$ for $P$ cross-validation parameters~\cite{lines2015time, bagnall2017great, lucas2019proximity}. Furthermore, since EE is an ensemble of 1-NN models, the classification time for each time series is also high -- $O(n \cdot L ^2)$. EE was the overall most accurate similarity or distance-based classifier on the UCR benchmark until PF~\cite{bagnall2017great}. EE was also used as a component of HIVE-COTE.

Next, we present our novel multivariate similarity \reviewtwo{and distance}-based ensemble Multivariate Elastic Ensemble (MEE). We keep the design of our multivariate ensemble similar to the univariate EE, except that MEE uses the multivariate similarity measures. 
Similar to EE, MEE also uses leave-one-out cross-validation of 100 parameters when choosing the parameters for similarity \reviewtwo{and distance} measures. 
Both EE and MEE also predict the final label of a test instance by using highest class probability. The class probability of each measure is weighted by the leave-one-out cross-validation accuracy of each measure on the training set.
Any ties are broken using a uniform random choice. Similarly to the original $DTW_I$ and $DTW_D$~\citep{shokoohi2017generalizing}, all measures used in MEE use all dimensions in the dataset.

We explore four variations of MEE, which are constructed as follows.

\begin{itemize}
    \item $\mathit{MEE_I}$ : An ensemble of eleven 1-NN classifiers formed using only independent multivariate similarity \reviewtwo{and distance} measures.
    \item $\mathit{MEE_D}$ : An ensemble of eleven 1-NN classifiers formed using only dependent multivariate similarity \reviewtwo{and distance} measures.
    \item $\mathit{MEE_{ID}}$ : An ensemble of 22 1-NN classifiers formed using eleven independent measures and eleven dependent measures.
    \item $\mathit{MEE_{A}}$ : An ensemble of eleven 1-NN classifiers formed by selecting either independent or dependent version of the measure based on its accuracy on the training set (ties are broken randomly).
\end{itemize}


}

\section{Experiments}
\label{sec:experiments}

    \reviewone{First we conduct experiments using 1-NN classifiers with single multivariate measures to investigate two hypotheses and then we present the experiments conducted with Multivariate Elastic Ensemble. Finally, we investigate the use of normalization and the runtime.}
    
    The first \reviewone{hypothesis} is that there are different datasets to which each of the new multivariate distance measures is best suited. 
    The second arises from the observation that there are datasets for which either the independent or dependent version of DTW are consistently more accurate than the alternative \cite{shokoohi2017generalizing}. However, it is not clear whether this is a result of there being an advantage in treating multivariate series as either a single series of multivariate points or multiple independent series of univariate points; or rather due to some other property of the measures. 
    
    It is credible that there should be some time series data for which it is beneficial to treat multiple variables as multivariate points in a single series. Suppose, for example, that the variables each represent the throughput of independent parts of a process and the quantity relevant to classification is aggregate throughput.  In this case, the sum of the values at each point is the relevant quantity. In contrast, if classification relates to a failure in any of those parts, it seems clear that independent consideration of each is the better approach.
    
    We seek to assess whether there are multivariate datasets for which each of dependent and independent analysis are best suited, or whether there are other reasons, such as their mathematical properties, that underlie the systematic advantage on specific datasets of either $DTW_I$ or $DTW_D$.
    
    
    We start by describing our experimental setup and the datasets we used.
    We then conduct an analysis of similarity \reviewtwo{and distance} measures in the context of TSC by comparing accuracy measures of independent and dependent versions.
    We then conduct a statistical test to determine if there is a difference between independent and
dependent versions of the measures.

    \subsection{Experimental Setup}
    \label{subsec:setup}

    We implemented a multi-threaded version of the multivariate similarity \reviewtwo{and distance} measures in Java.
    We also release the full source code in the github repository: \url{https://github.com/dotnet54/multivariate-measures}.

    In these experiments, for parameterization of the measures, we use leave-one-out cross-validation of 100 parameters for each similarity \reviewtwo{and distance} measure. We follow the same settings proposed in~\cite{lines2015time}. This parameterization is also used in HIVE-COTE, PF, and TS-CHIEF.


    In this study, we use multivariate datasets obtained from \url{https://www.timeseriesclassification.com}.
    \reviewone{For each dataset, we use 10 resamples for training with a train/test split ratio similar to the default train/test split ratio provided in the repository.}  
    Out of the available 30 datasets, we use 23 datasets in this study.
    Since we focus only on fixed-length datasets, the four variable length datasets (\emph{CharacterTrajectories}, \emph{InsectWingbeat}, \emph{JapaneseVowels}, and
\emph{SpokenArabicDigits}) are excluded from this study.
    We also omit \emph{EigenWorms}, \emph{MotorImagery} and, \emph{FaceDetection}, which take
too long to run the leave-one-out cross-validation for 100 parameters in a practical time frame.
    \mbox{Table~\ref{tbl:datasets}} (on~page~\pageref{tbl:datasets}) summarizes the characteristics of the 23 fixed length datasets. The main results are obtained for non-normalized datasets, except for four datasets that are already normalized in the archive. We explore the influence of the normalization in Section~\ref{subsec:norm-vs-unnorm}.
    Further descriptions of each dataset can be found in~\cite{bagnall2018uea}.

    \subsection{Accuracy of Independent \textit{Vs} Dependent Measures}
    \label{subsec:indep-v-dep}

    First, we look at the accuracy of each measure used with a 1-NN classifier.
    Tables~\ref{tbl:acc-indep} and~\ref{tbl:acc-dep} (on~page~\pageref{tbl:acc-indep})
    present the accuracy for independent measures and
    dependent measures, respectively. For each dataset, the highest accuracy is typeset in bold.
    Of the values reported in Tables~\ref{tbl:acc-indep} and~\ref{tbl:acc-dep}, accuracy for measures other
    than
    Euclidean distance (labeled ``$L_2$'' in the table) and DTW are newly published results in this paper. 
\reviewone{
Our first observation is that for every similarity \reviewtwo{and distance} measure, except $\mathit{LCSS_D}$, there is at least one dataset for which that measure obtains the highest accuracy. This is consistent with our first hypothesis, that each measure will have datasets for which it is well suited.
}

To compare multiple algorithms over the multiple datasets, first a Friedman test is performed to reject the null hypothesis.
The null hypothesis is that there is no significant difference in the mean ranks of
the multiple algorithms (at a statistical significance level $\alpha=0.05$).
In cases where the null-hypothesis~\reviewone{of the Friedman test} is rejected, we use the Wilcoxon signed rank test to compare the pair-wise
difference in ranks between algorithms, and then use Holm--Bonferroni's method to adjust for family-wise errors~\cite{demvsar2006statistical, benavoli2016should}.

\begin{figure}[ht]
        \centering
        \includegraphics[width=.8\linewidth]{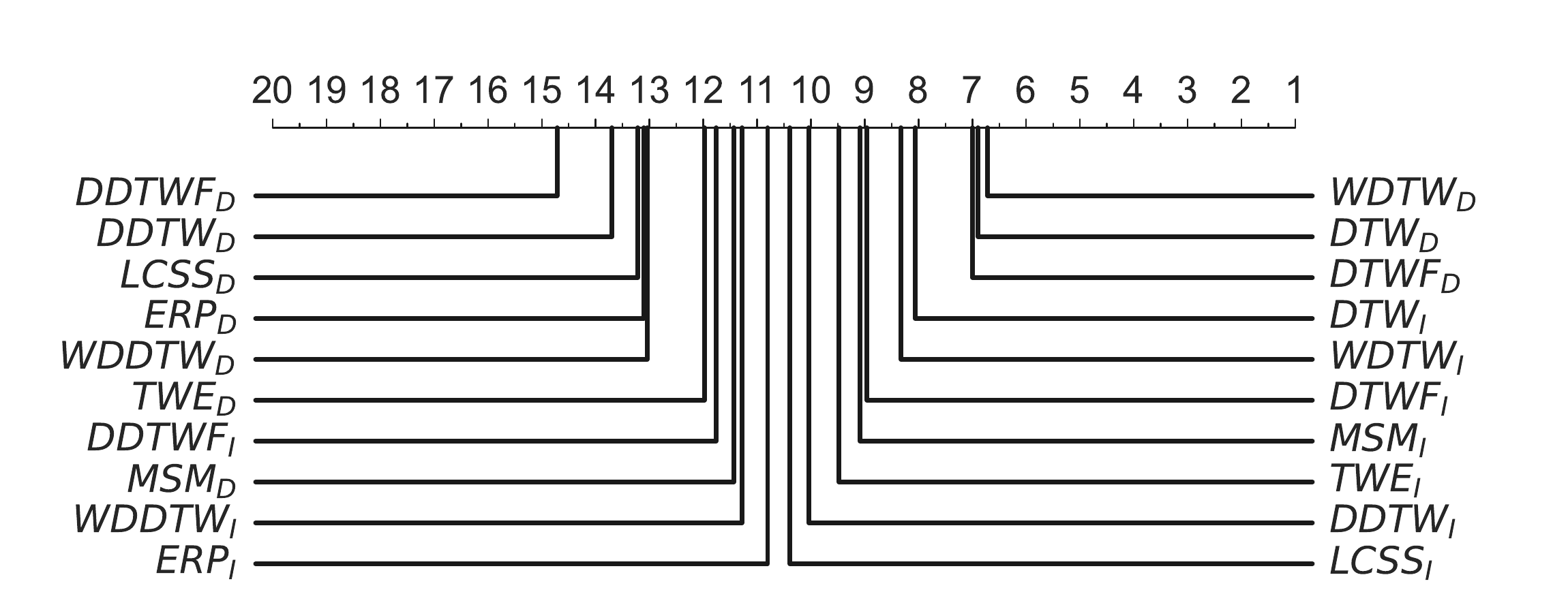}
        \caption{\reviewone{Average accuracy ranking diagram showing the ranks of the measures on the error rates (thus
more accurate measures are to the right side). For clarity, we have removed the horizontal lines (``cliques'') that normally connect groups of classifiers that are statistically indistinguishable.}}
        \label{fig:cd-measures}
    \end{figure}

\reviewone{Figure~\ref{fig:cd-measures} displays mean ranks (on error) between all similarity \reviewtwo{and distance} measures.} Measures on the right side indicate higher rank in accuracy (lower error).
We do not include $L2$ distance here to we focus on ``elastic'' measures only.
Since we use Holm--Bonferroni's correction, there is not a single ``critical difference value'' that applies to all pairwise comparisons.
Hence, we refer to these visualizations as ``average accuracy ranking diagrams''.
\reviewone{
For clarity, we also remove the horizontal lines that group statistically different classifiers. We refer to the p-value tables and highlight the statistically indistinguishable pairs most relevant to the discussion in the text. 
}

In Figure~\ref{fig:cd-measures}, $\mathit{WDTW_D}$, which is to the further right, is the most accurate measure on
the evaluated datasets.
$\mathit{WDTW_D}$ obtained a ranking of 6.7174 from the Wilcoxon test.
By contrast, $\mathit{DDTWF_D}$ (ranked 14.7174) is the least accurate measure on these datasets. 
After Holm--Bonferroni's correction, computed p-values indicate there are five pairs that are statistically different from each other. They are: $\mathit{DTW_D}$ and $\mathit{ERP_D}$, $\mathit{ERP_D}$ and $\mathit{WDTW_D}$, $\mathit{DDTWF_D}$ and $\mathit{DTW_D}$, $\mathit{DDTWF_D}$ and $\mathit{WDTW_D}$, $\mathit{LCSS_D}$ and $\mathit{WDTW_D}$.

    \subsection{Are Independent and Dependent Measures Significantly Different?}
    \label{subsec:indep-v-dep-heatmap}

    In this section, we test if there are datasets for which independent or dependent version is always more accurate.
    We also test if there is a statistically significant difference between
independent and dependent similarity \reviewtwo{and distance} measures.
    Answering these questions will help us determine the usefulness of developing these two variations of the
multivariate similarity \reviewtwo{and distance} measures.
    It will also help us to construct ensembles of similarity \reviewtwo{and distance} measures with more diversity, that is expected
to perform well in terms of accuracy over a wide variety of datasets.

\begin{figure*}[ht]
        \centering
        \includegraphics[width=0.9\linewidth]{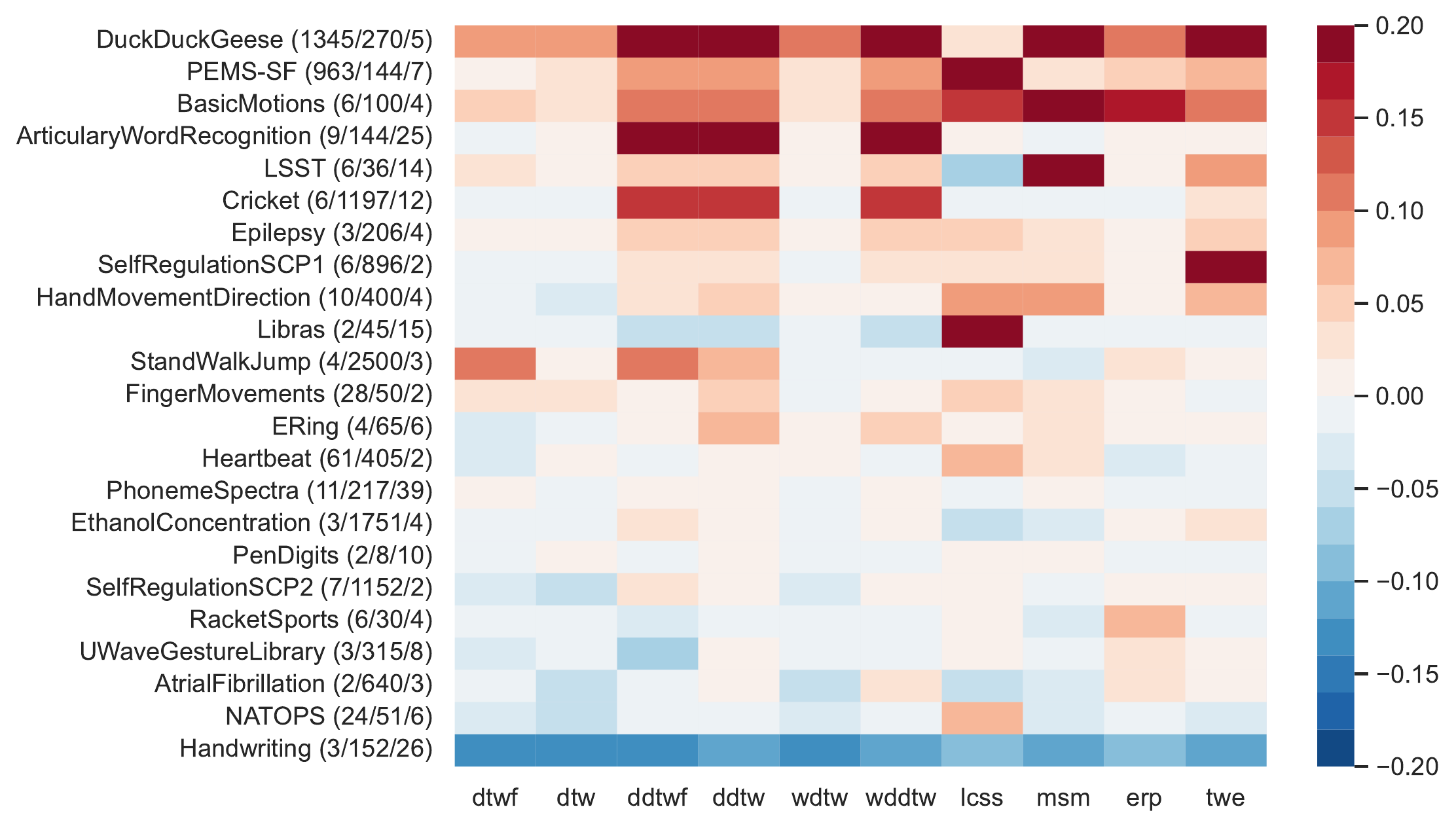}
        \caption{Heatmap showing the difference in accuracy between independent and dependent versions of the measures - deeper reddish colours indicate cases where independent is more accurate (positive on the scale), and deeper bluish colours indicate cases where dependent is more accurate (negative on the scale).
        The datasets are sorted based on average colour values to show contrasting colours on the two ends.
(Dimensions / length / number of classes are shown in the bracket after the dataset name.)
        }
        \label{fig:heatmap}
    \end{figure*}

    Figure~\ref{fig:heatmap} shows the difference in accuracy between independent and dependent versions of
the measures - deeper reddish colours indicate cases where independent is more accurate (positive on the scale), and deeper bluish colours indicate cases where dependent is more accurate (negative on the scale).
The datasets are sorted based on average colour values to show contrasting colours on the two ends.
Dimensions $D$, length $L$, number of classes $c$ are given in the bracket after the dataset name.

    From Figure~\ref{fig:heatmap} we observe that there are datasets for which either independent or dependent is
always more accurate.
    For example, the independent versions of all measures are consistently more accurate for datasets \emph{DuckDuckGeese}, \emph{PEMS-SF} and \emph{BasicMotions} (indicated by red colour rows in the heatmap).
    On the other hand, we see that \emph{Handwriting} always wins for the dependent versions (indicated by the
blue colour row).

Next we statistically investigate the hypothesis that there are some multivariate TSC tasks that are inherently best suited to either treating the multivariate series as a single series of multvariate points or as multiple independent series of univariate points. 
To this end we present the results of a Wilcoxon signed-rank test on each of the 10 pairs of
measures (without $L_2$), to test whether the difference between accuracy of independent and dependent versions across 23 datasets are statistically significant.
We conduct this test with the null hypothesis that the mean of the difference between the accuracy of the
independent and dependent versions will be zero.
We reject the null hypothesis with statistical significance value $\alpha=0.05$, and accept that there is a significant statistical difference in accuracy if
the $p\leq\alpha$.
Table~\ref{tbl:wilcoxon} shows the $p$-value for each dataset. The bold values mark the $p$-values for which there is a significant difference.
We also report the adjusted $\alpha$ value after Holm--Bonferroni corrections, $\alpha_{HB}$.
Before Holm--Bonferroni correction, out of the 23 datasets, we observe that for 8 datasets there is a statistically
significant difference in accuracy between independent and dependent measures.
After Holm--Bonferroni correction, we still find 5 statistically significant differences (where $p\leq\alpha_{HB}$) and so conclude there are indeed datasets that are inherently best suited to either independent or dependent treatment.

\begin{table}[ht]
\centering
\caption{p-values for two-sided Signed Rank Wilcoxon test with $\alpha=0.05$ (significant values are in bold face) and $\alpha$ values adjusted for multiple testing using the Holm-Bonferroni correction.}
\label{tbl:wilcoxon}
\begin{tabular}{lrr}
\toprule
                  dataset &          p-value & $\alpha_{HB}$ \\
\midrule
             BasicMotions &  \textbf{0.0020} &                        \textbf{0.0021} \\
            DuckDuckGeese &  \textbf{0.0020} &                        \textbf{0.0022} \\
                 Epilepsy &  \textbf{0.0020} &                        \textbf{0.0023} \\
                  PEMS-SF &  \textbf{0.0020} &                        \textbf{0.0024} \\
              Handwriting &  \textbf{0.0020} &                        \textbf{0.0025} \\
          FingerMovements &  \textbf{0.0249} &                        0.0026 \\
ArticularyWordRecognition &  \textbf{0.0371} &                        0.0028 \\
                     LSST &  \textbf{0.0488} &                        0.0029 \\
    HandMovementDirection &  0.0528 &                        0.0031 \\
       SelfRegulationSCP1 &  0.0645 &                        0.0033 \\
                    ERing &  0.0645 &                        0.0036 \\
                   NATOPS &  0.0840 &                        0.0042 \\
                   Libras &  0.0840 &                        0.0038 \\
             RacketSports &  0.1934 &                        0.0045 \\
            StandWalkJump &  0.3081 &                        0.0050 \\
      UWaveGestureLibrary &  0.3750 &                        0.0056 \\
       AtrialFibrillation &  0.3750 &                        0.0063 \\
                PenDigits &  0.4316 &                        0.0071 \\
                  Cricket &  0.5566 &                        0.0083 \\
           PhonemeSpectra &  0.6953 &                        0.0100 \\
       SelfRegulationSCP2 &  0.8457 &                        0.0125 \\
     EthanolConcentration &  1.0000 &                        0.0167 \\
                Heartbeat &  1.0000 &                        0.0250 \\
\bottomrule
\end{tabular}
\end{table}

\reviewone{This finding leads to the questions of why this is the case and whether it is predictable which strategy will prevail.

In general, we observe that the dependent strategy tends to perform poorly when there are a large number of dimensions in the dataset. The two datasets for which the independent strategy performs most strongly are the only two with more than 100 dimensions. The explanation for this may simply be that high dimensional datasets suffer from the curse of dimensionality and that the $L_2$-norms between different sets of high dimensional points carry little information.

For lower dimensional data, we hypothesize that the independent strategy will be effective when the interactions between the dimensions carry little information about the classification task and the dependent strategy will be effective when the interactions between the dimensions carry much information about the classification task.

Consider the \emph{Handwriting} dataset for which all dependent measures are more accurate than their corresponding independent measures (see Figure~\ref{fig:heatmap}). It has three accelerometer values recording hand written letters. It seems intuitive that to distinguish a straight line from curved; horizontal from vertical; and writing from repositioning the pen; it is necessary to consider all three accelerometers together.  To test this we compare the accuracy of 1-NN classifiers using each dimension alone to the accuracy of the independent and dependent strategies. The results are shown in Table~\ref{tbl:case_study_dims}. Firstly, we observe that for \emph{Handwriting} dataset, $DTW_D$ performs more accurately than $DTW_I$ (0.61~\emph{versus}~0.46). In addition, we see that using $DTW_D$ with all three dimensions of the dataset is more accurate than using only a single dimension (last three rows). This shows that dependent measures can be effective for datasets that contain information in the interactions between dimensions.

We contrast this to the six dimensional \emph{BasicMotions} dataset for which all independent measures are more accurate than their corresponding dependent measures (see Figure~\ref{fig:heatmap}).  Comparing the accuracy of the single dimension 1-NN classifiers to that of the independent and dependent strategies, we see that each of three of the dimensions when used alone can attain 100\% accuracy. When all dimensions are considered together, $DTW_I$ also performs better than $DTW_D$ (1.00~\emph{versus}~0.96). 
This suggests that independent measures will have good performance when individual dimensions carry substantial information about the class without need to consider interdependencies between dimensions.
}

\reviewone{

\begin{table}[ht]
\centering
\caption{The table shows the accuracy of two datasets with two independent DTW and dependent DTW when all dimensions are used and a single dimension is used. \emph{BasicMotion} was selected because it performs more accurately on all independent measures, and \emph{Handwriting} was selected because it performs more accurately on all dependent measures (see Figure~\ref{fig:heatmap}). }
\label{tbl:case_study_dims}
\begin{tabular}{rllll} 
\toprule
                    Dataset & Dimensions & $DTW_I$ & $DTW_D$ & DTW \\
\midrule
Handwriting & 3 & 0.46 & 0.61 & \\ [0.5em]
\midrule
Handwriting & 1 (dim \#0) &  & & 0.32 \\ [0.5em]
Handwriting & 1 (dim \#1) &  & & 0.24 \\  [0.5em]
Handwriting & 1 (dim \#2) &  & & 0.35 \\ [0.5em]
\midrule
BasicMotion & 6 & 1.00 & 0.96 &  \\ [0.5em]
\midrule
BasicMotion & 1 (dim \#0) &  &  & 1.00 \\ [0.5em]
BasicMotion & 1 (dim \#1) &  &  & 1.00 \\ [0.5em]
BasicMotion & 1 (dim \#2) &  & & 0.90 \\ [0.5em]
BasicMotion & 1 (dim \#3) &  & & 0.98 \\ [0.5em]
BasicMotion & 1 (dim \#4) &  & & 0.85 \\ [0.5em]
BasicMotion & 1 (dim \#5) &  & & 1.00 \\ [0.5em]
\bottomrule
\end{tabular}
\end{table}

}


It is interesting to note that there appears to be considerable correlation between the relative desirability of the independent and dependent approaches across all the measures that are applied to the derivative of the original series, $DDTWF$, $DDTW$ and $WDDTW$. It particularly stands out that there seems to be a strong advantage to independent variants of these measures with respect to \emph{ArticularyWordRecognition}, \emph{DuckDuckGeese} and \emph{Cricket}. Possible connections between transformations and the relative efficacy of independent or dependent approaches may be a productive topic for future research.

\reviewone{

\subsection{Multivariate Measures \emph{vs} MEE}
\label{ch6:subsec:measures-vs-mee}

EE showed that 1-NN ensembles formed using a diverse set of similarity \reviewtwo{and distance} measures are more accurate than any of the single measures on the 85 univariate UCR datasets~\citep{lines2015time, bagnall2017great}. In this experiment, we replicate this evaluation in the multivariate context. We compare single independent and dependent measures with MEE ensembles to assess whether each of the three versions of MEE are significantly different to the individual measures.

Figures~\ref{fig:ch6-cd-measures-mee-indep} and~\ref{fig:ch6-cd-measures-mee-dep} show accuracy rankings of independent measures and $\mathit{MEE_I}$; and dependent measures and $\mathit{MEE_D}$, respectively. 
Figure~\ref{fig:ch6-cd-measures-mee-indep} indicates that $\mathit{MEE_I}$ obtains the best accuracy. However, an investigation of the $p$-values indicate that $\mathit{MEE_I}$ is not significantly different from seven independent measures ($\mathit{WDTW_I}$, $\mathit{DTW_I}$, $\mathit{DTWF_I}$, $\mathit{MSM_I}$, $\mathit{ERP_I}$, $\mathit{WDDTW_I}$, and $\mathit{DDTWF_I}$).
As for, Figure~\ref{fig:ch6-cd-measures-mee-dep}, the computed p-values indicates that $\mathit{MEE_D}$ is significantly different to individual measures except for $\mathit{WDTW_D}$,$DTWF_D$ and $DTW_D$.
Both results indicate that $\mathit{MEE_I}$ and $\mathit{MEE_D}$ are more accurate than individual measures with 1-NN. 

\begin{figure}[]
    \centering
    \includegraphics[width=.8\linewidth]{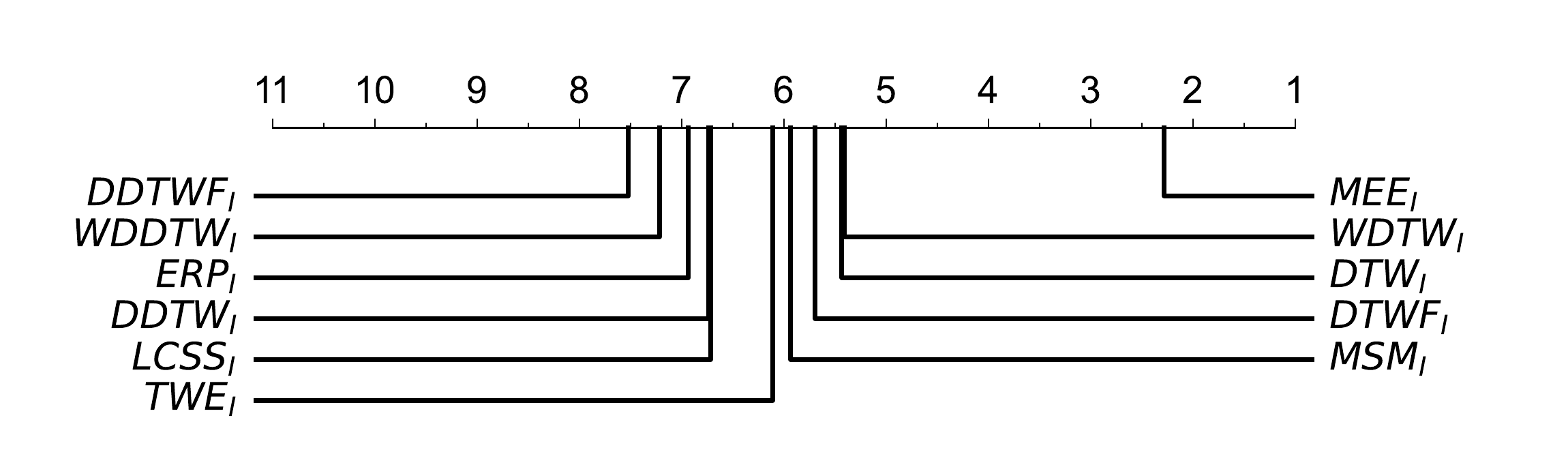}
    \caption{Average accuracy ranking diagram showing the ranks on the error rate of the independent similarity \reviewtwo{and distance} measures and $\mathit{MEE_I}$.}
    \label{fig:ch6-cd-measures-mee-indep}
\end{figure}

\begin{figure}[]
    \centering
    \includegraphics[width=.8\linewidth]{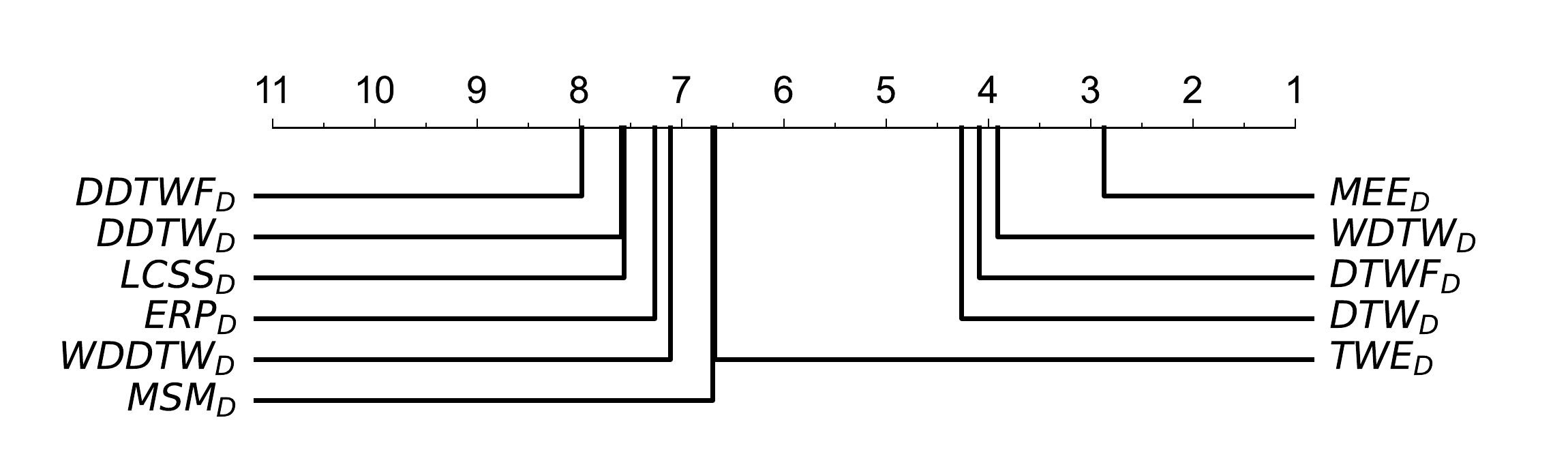}
    \caption{Average accuracy ranking diagram showing the ranks on the error rate of the dependent similarity \reviewtwo{and distance} measures and $\mathit{MEE_D}$.}
    \label{fig:ch6-cd-measures-mee-dep}
\end{figure}

Figure~\ref{fig:ch6-cd-measures-mee} shows accuracy ranking of our four ensembles presented in Section~\ref{subsec:mee} \textit{versus} the top five (out of twenty-one) individual similarity \reviewtwo{and distance} measures with 1-NN.
We can observe that all ensembles are more accurate than the classifiers using a single measure.
We found that the $\mathit{MEE_{A}}$ (avg. rank 2.5870) performs better than $\mathit{MEE_{ID}}$ (avg. rank 2.9130), $\mathit{MEE_I}$ (avg. rank 3.7174) and $\mathit{MEE_D}$ (avg. rank 5.0435). The computed p-values show that the difference between $\mathit{MEE_{A}}$ and $\mathit{MEE_D}$ is statistically significant.
Based on these results we select $\mathit{MEE_{A}}$ as the final design of $\mathit{MEE}$. 

\begin{figure}[]
    \centering
    \includegraphics[width=.8\linewidth]{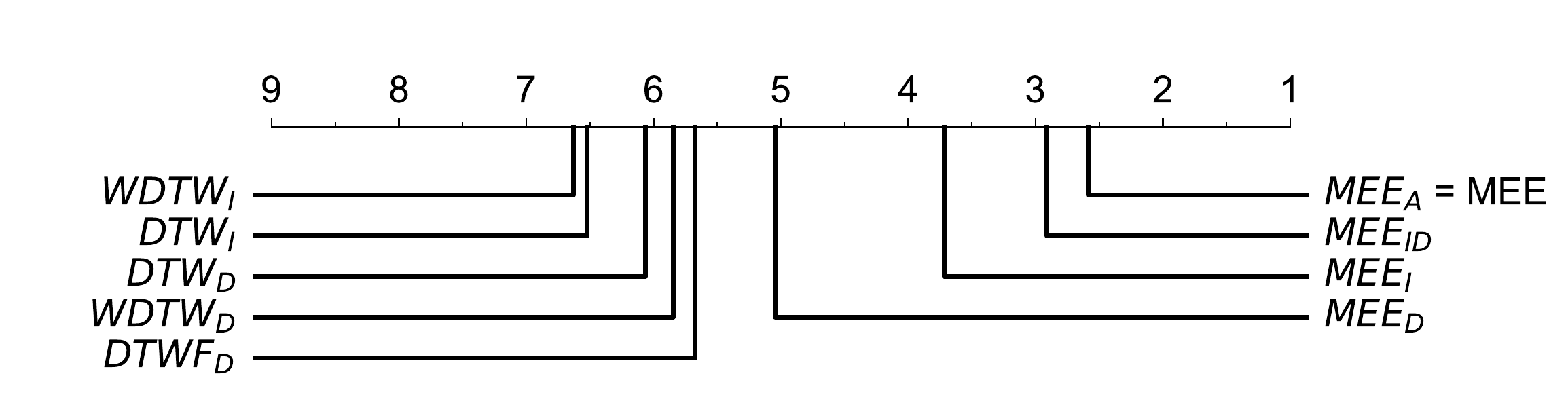}
    \caption{Average accuracy ranking diagram showing the ranks on the error rate of the top five similarity \reviewtwo{and distance} measures and four variants of MEE.}
    \label{fig:ch6-cd-measures-mee}
\end{figure}


}

\reviewone{

\subsection{MEE \emph{vs} SOTA Multivariate TSC Algorithms}
\label{ch6:subsec:mee-vs-sota}

Next, we compare $\mathit{MEE}$ (\ie $\mathit{MEE_A}$) with six state-of-the-art multivariate TSC algorithms~\citep{ruiz2021great, middlehurst2021hive}. Figure~\ref{fig:ch6-cd-sota-mve-ruiz-hc2} shows the accuracy ranking of our ensembles and these algorithms.

\begin{figure}
    \centering
    \includegraphics[width=.8\linewidth]{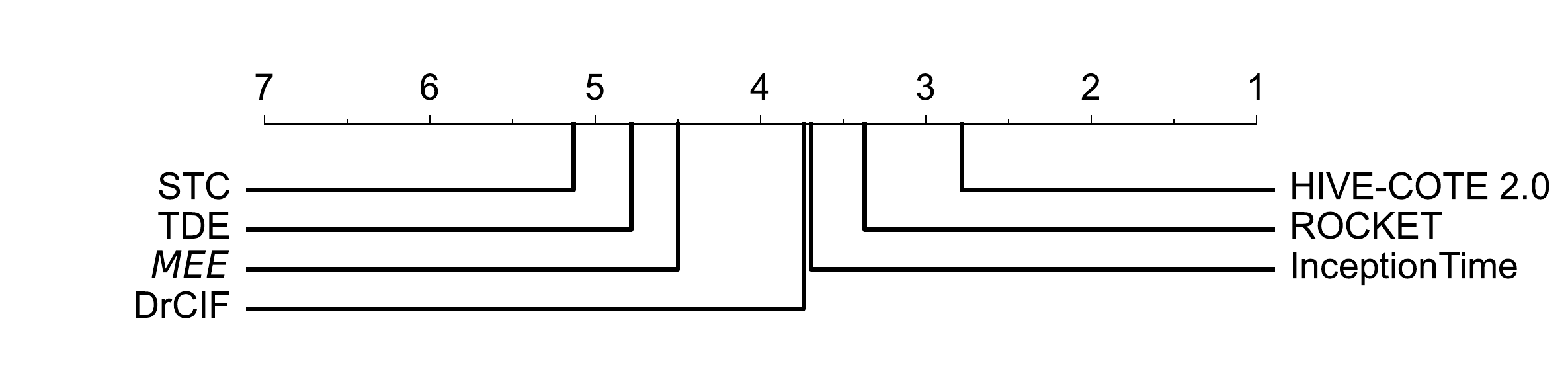}
    \caption{Average accuracy ranking diagram showing the ranks on the error rate of six classifiers from Ruiz~et~al.~\citep{ruiz2021great} and Middlehurst~et~al.~\citep{middlehurst2021hive} and our best performing ensemble, $\mathit{MEE}$.}
    \label{fig:ch6-cd-sota-mve-ruiz-hc2}
\end{figure}

The most accurate algorithm, HIVE-COTE 2.0, obtained an average rank of 2.7826. It is followed by ROCKET (ranked 3.3696), InceptionTime (3.6957), DrCIF (ranked 3.7391) and then our ensemble $\mathit{MEE}$ (4.5000). Our ensemble $\mathit{MEE}$ is not significantly different to more complex leading classifiers such as HIVE-COTE 2.0, ROCKET and InceptionTime. 
Based on the p-values, only 2 pairs of classifiers are significantly different. They are: STC \emph{vs} HIVE-COTE~2.0~and TDE~\emph{vs}~HIVE-COTE~2.0. 


Finally, Table~\ref{tbl:sota-mee-accuracy} shows the accuracy of leading algorithms (selected from Figure~\ref{fig:ch6-cd-sota-mve-ruiz-hc2}) and our ensembles. Across the all algorithms, $\mathit{MEE}$ achieves the highest accuracy for 2 datasets. By comparison, shapelet-based STC is highest on 1 dataset and dictionary-based TDE on 2 datasets. Interestingly, highly ranked ROCKET has highest accuracy on only two datasets while InceptionTime and HIVE-COTE 2.0 have highest accuracy six and eight times, respectively.

\begin{table}
\centering
\caption{Accuracy of our ensemble $MEE$ compared against top multivariate TSC algorithms on 23 datasets from UEA Multivariate TS Archive.                   Column names are shortened as follows: IT for InceptionTime, RT for ROCKET, HC2 for HIVE-COTE 2.0.                   Wins indicate the number of time each classifier achieved the highest accuracy for each dataset.}
\label{tbl:sota-mee-accuracy}
\begin{tabular}{p{0.6cm}*{16}{p{0.65cm}}}
\toprule
dataset &     $MEE_{I}$ &     $MEE_{D}$ &    $MEE_{ID}$ &         $MEE$ &         DrCIF &           TDE &            IT &           STC &            RT &           HC2 \\
\midrule
    AWR &          0.99 &          0.99 &          0.99 &          0.99 &          0.98 &          0.98 &          0.99 &          0.98 &          1.00 & \textbf{1.00} \\
     AF &          0.27 & \textbf{0.35} &          0.32 &          0.31 &          0.23 &          0.30 &          0.22 &          0.32 &          0.25 &          0.28 \\
     BM & \textbf{1.00} &          0.91 &          0.97 & \textbf{1.00} & \textbf{1.00} &          0.99 & \textbf{1.00} &          0.98 &          0.99 &          0.99 \\
     CR &          1.00 & \textbf{1.00} & \textbf{1.00} & \textbf{1.00} &          0.99 &          0.99 &          0.99 &          0.99 & \textbf{1.00} &          1.00 \\
    DDG &          0.58 &          0.47 &          0.58 &          0.60 &          0.58 &          0.32 & \textbf{0.63} &          0.43 &          0.46 &          0.50 \\
     EP &          0.98 &          0.96 &          0.97 &          0.98 &          0.99 &          1.00 &          0.99 &          0.99 &          0.99 & \textbf{1.00} \\
     EC &          0.30 &          0.31 &          0.30 &          0.31 &          0.67 &          0.53 &          0.28 & \textbf{0.82} &          0.45 &          0.79 \\
     ER &          0.97 &          0.95 &          0.97 &          0.96 &          0.98 &          0.94 &          0.92 &          0.84 &          0.98 & \textbf{0.99} \\
     FM & \textbf{0.57} &          0.54 &          0.56 &          0.56 &          0.56 &          0.54 &          0.56 &          0.53 &          0.55 &          0.55 \\
    HMD &          0.34 &          0.29 &          0.34 &          0.34 & \textbf{0.46} &          0.38 &          0.42 &          0.35 &          0.45 &          0.40 \\
     HW &          0.49 &          0.60 &          0.58 &          0.60 &          0.34 &          0.56 & \textbf{0.66} &          0.29 &          0.57 &          0.56 \\
     HB &          0.73 &          0.72 &          0.73 &          0.72 & \textbf{0.76} &          0.72 &          0.73 &          0.72 &          0.72 &          0.73 \\
    LIB &          0.88 &          0.89 &          0.89 &          0.89 &          0.91 &          0.88 &          0.89 &          0.84 &          0.91 & \textbf{0.93} \\
   LSST &          0.58 &          0.55 &          0.59 &          0.58 &          0.55 &          0.56 &          0.34 &          0.58 &          0.63 & \textbf{0.64} \\
   NATO &          0.79 &          0.83 &          0.81 &          0.81 &          0.84 &          0.82 & \textbf{0.97} &          0.84 &          0.89 &          0.89 \\
     PD &          0.99 &          0.99 &          0.99 &          0.99 &          0.99 &          0.97 & \textbf{1.00} &          0.98 &          1.00 &          1.00 \\
   PEMS &          0.82 &          0.78 &          0.81 &          0.82 &          1.00 & \textbf{1.00} &          0.83 &          0.98 &          0.86 &          1.00 \\
     PS &          0.19 &          0.19 &          0.19 &          0.19 &          0.31 &          0.23 & \textbf{0.37} &          0.31 &          0.28 &          0.29 \\
     RS &          0.87 &          0.87 &          0.88 &          0.87 &          0.90 &          0.89 &          0.92 &          0.88 &          0.93 & \textbf{0.93} \\
   SRS1 &          0.83 &          0.82 &          0.83 &          0.83 &          0.87 &          0.83 &          0.85 &          0.85 &          0.87 & \textbf{0.88} \\
   SRS2 &          0.51 &          0.51 &          0.52 &          0.51 &          0.51 & \textbf{0.53} &          0.52 &          0.52 &          0.51 &          0.50 \\
    SWJ &          0.35 &          0.30 &          0.31 &          0.33 &          0.41 &          0.36 &          0.42 &          0.44 & \textbf{0.46} &          0.44 \\
     UW &          0.92 &          0.93 &          0.93 &          0.93 &          0.92 &          0.93 &          0.91 &          0.87 &          0.94 & \textbf{0.95} \\
     \bottomrule
   Wins &             2 &             2 &             1 &             2 &             3 &             2 &             6 &             1 &             2 &             8 \\
\bottomrule
\end{tabular}
\end{table}


}

\reviewone{
\subsection{Performance on Normalized vs Unnormalized data}
\label{subsec:norm-vs-unnorm}

We ran experiments for all measures for both z-normalized and unnormalized datasets. Note that 4 datasets --- \emph{ArticularyWordRecognition}, \emph{Cricket}, \emph{HandMovementDirection}, and \emph{UWaveGestureLibrary} --- are already normalized.  These were excluded from this study as it is not possible to derive unnormalized versions.
  We z-normalized each of the remaining datasets on a per series, per dimension basis. We found that the accuracy is higher without normalization. This agrees with a recent paper which conducted a similar experiment using $DTW_I$ and $DTW_D$ \cite{ruiz2021great}. 

We include the results for all measures in our github repository\footnote{\gitrepo}. To summarize the results, Figure~\ref{fig:normalize-vs-unnormalized} shows a scatter plot comparing the accuracy of MEE with normalized and unnormalized data. This comparison was conducted on the default train and test split and excludes the already  normalized four datasets in the archive. The results show MEE with unnormalized datasets win 14 times and loses 4 times with 1 tie. In addition, for both $\mathit{MEE_I}$ (12/2/5 win/draw/loss)  and $\mathit{MEE_D}$ (11/1/7 win/draw/loss) we also observed that unnormalized data works better. This is why we used  unnormalized datasets for all other experiments in the paper.

\begin{figure}[!h]
    \centering
    \includegraphics[width=.6\linewidth]{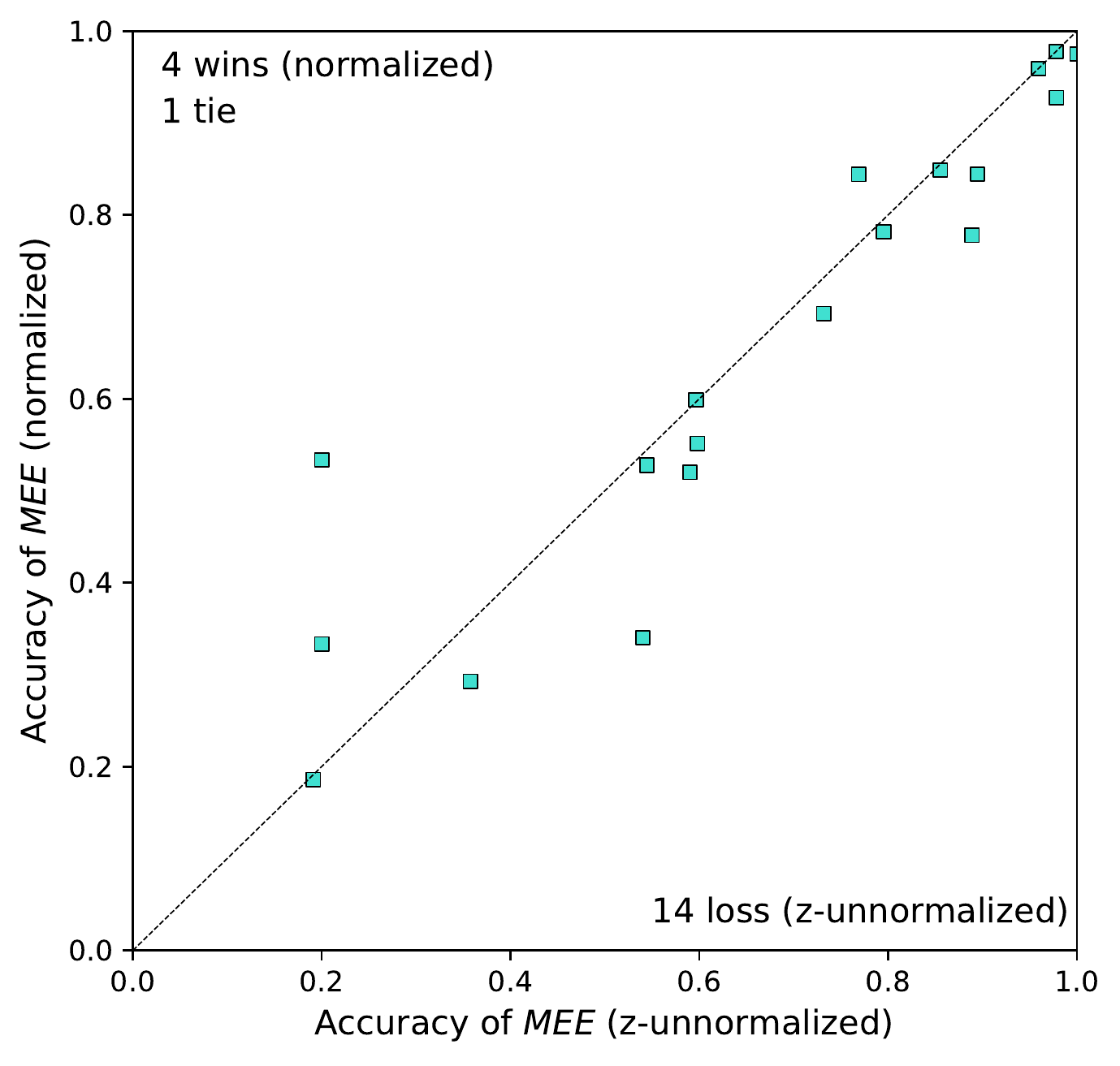}
    \caption{Accuracy comparison of MEE with unnormalized data \emph{versus} z-normalized data. This comparison excludes the already normalized four datasets in the archive.}
    \label{fig:normalize-vs-unnormalized}
\end{figure}

    However, we note that this does not indicate that not normalizing is always the optimal solution for all datasets.
    Sometimes normalization can be useful when using similarity \reviewtwo{and distance} measures.
    For example, consider a scenario with two dimensions temperature (e.g a scale from 0 to 100 degree Celsius) and relative humidity as a proportion (between 0 and 1).
    In such a case, temperature will dominate the result of the similarity or distance calculation, and normalization will help to compute the similarity or distance with similar scales across the dimensions.

}

\reviewone{
\subsection{Runtime}
\label{subsec:runtime}

In this section, we provide a summary of the runtime information and discuss the time complexity of the multivariate measures and the MEE.

We ran the experiments on a cluster of Intel(R) Xeon(R) CPU E5-2680 v3 @ 2.50 GHz CPUs, using 16-threads. The total time to train 23 datasets with leave-one-out cross-validation with 10 resamples of the training set was about 10,423 hours (wall clock time).
On average, one fold of leave-one-out cross-validation required 1,042 hours of train time. 
The two slowest datasets were \emph{PEMS-SF} (149 hours) and \emph{PhonemeSpectra} (650 hours) per fold, about 76\% of the total training time. Moreover, 11 out of 23 datasets took less than one hour per dataset to train. 

The slowest measure to train was $MSM_D$, which took a total of 3,553 hours across all datasets and the 10 folds (almost 35\% of the total training time). The second slowest measure $MSM_I$ took 1,109 hours and the third slowest $TWE_I$ took 851 hours. By contrast, the fastest measure to train was $DTWF_D$ and took just one hour. This is excluding Euclidean distance with no parameter which only took few minutes. Note that since MEE uses training accuracy to weight the ensemble predictions, leave-on-out cross-validation is performed at least once even if there are no parameters or a single parameter (e.g. DTWF). Figure~\ref{fig:runtime} shows the average runtime of each measure per fold.

\begin{figure}[h]
    \centering
    \includegraphics[width=.8\linewidth]{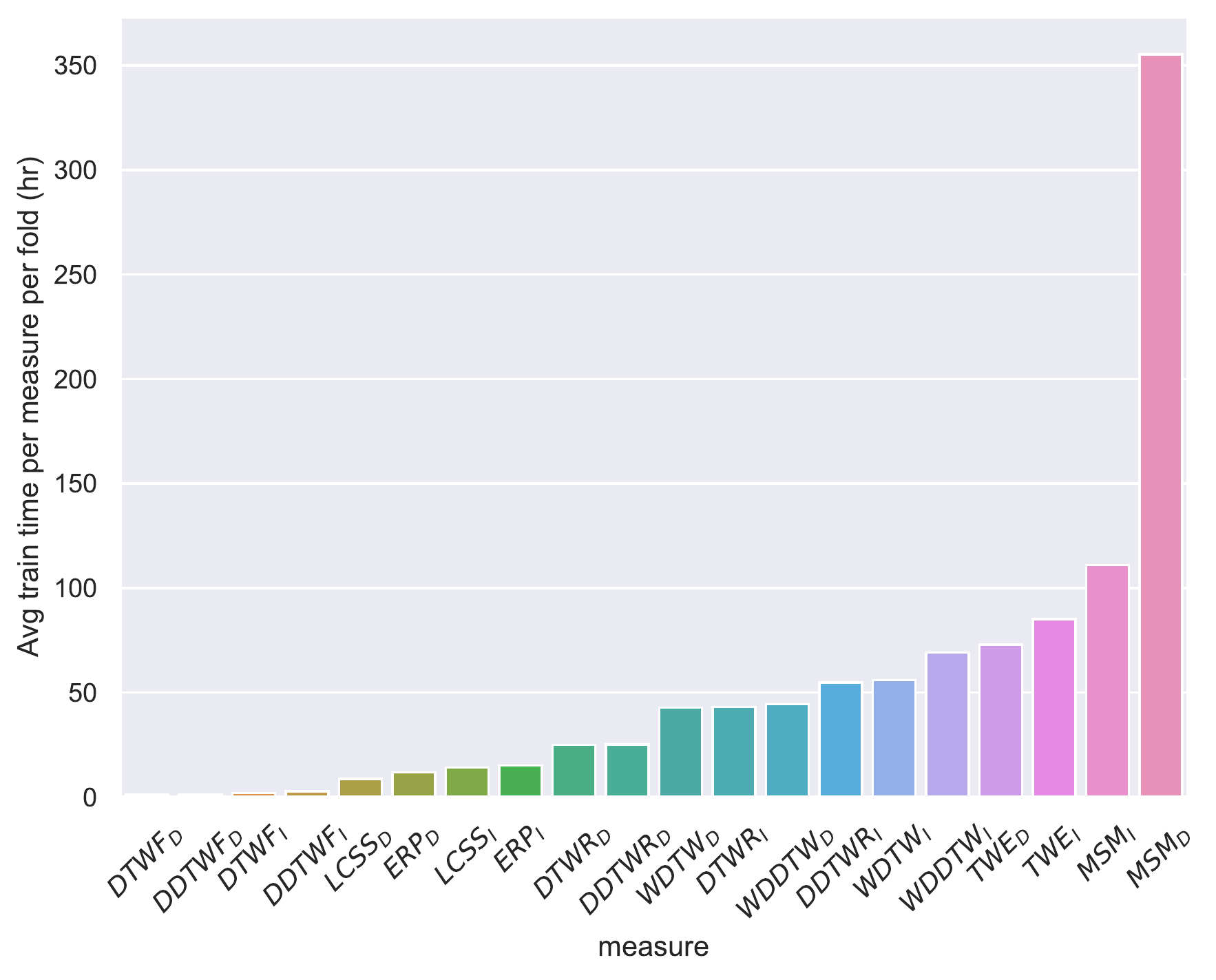}
    \caption{Average runtime of each measure per fold in hours.}
    \label{fig:runtime}
\end{figure}

Compared to classifiers such as ROCKET, our ensemble, MEE created from these measures have very high training time. However, the goal of MEE is not to tackle the scalability issue but to create a multivariate similarity \reviewtwo{and distance} based-classifier as a baseline for accuracy comparison with other similarity \reviewtwo{and distance}-based multivariate classifiers. This is similar to the univariate EE, which was extremely slow with a time complexity of $O(n^2 \cdot L^2 \cdot P)$ for $P$ cross-validation parameters~\cite{lines2015time, bagnall2017great, lucas2019proximity}. Despite its speed, EE stimulated much research in TSC and helped the development of more accurate classifiers such as the earlier versions of HIVE-COTE, and much faster classifiers such as PF and TS-CHIEF. Training time and test time complexity of MEE is: $O(n^2 \cdot L^2 \cdot D \cdot P)$ and $O(n \cdot L^2 \cdot D)$, respectively.

Measures such as DTW can be scaled to millions of time series when used in conjunction with lower bounding and early abandoning~\citep{keogh2006lb_keogh,ding2008querying, tan2017indexing}. Therefore, if various research on lower bounding and early abandoning of other similarity \reviewtwo{and distance} measures~\citep{keogh2009supporting,lemire2009faster,tan2019elastic,herrmann2021early} are combined and extended to multivariate measures, then it might be possible to create a more scalable version of MEE.

}

    \section{Conclusion}
    \label{sec:conclusion}

    In this paper, we present multivariate versions of seven commonly used elastic similarity \reviewtwo{and distance} measures.
    Our approach is inspired by independent and dependent DTW measures, which have proved very successful as strategies for extending univariate DTW to the multivariate case. 
    
    These measures can be used in a wide range of time series analysis tasks including classification, clustering, anomaly detection, indexing, subsequence search and segmentation. This study demonstrates their utility for time series classification.
    Our experiments show that each of the univariate similarity \reviewtwo{and distance} measures excels at nearest neighbor classification on different datasets, highlighting the importance of having a range of such measures in our analytic toolkits.
    
   They also show that there are datasets for which the independent version of DTW is more accurate than the dependent version and vice versa. Until now there was no way to determine whether this is a result of a fundamental difference between treating each dimension independently or not, or whether it arises from other properties of the algorithms. Our results showing that there are some datasets for which dependent or independent treatments are consistently superior across all distance measures provides strong support for the conclusion that it is a fundamental property of the datasets, that either the variables are best considered as a single multivariate point at each time step or are not. 
    
    We observe that the dependent strategy tends to perform poorly when there are a large number of dimensions in the dataset. Addressing this limitation  may be a productive direction for future research.  \reviewone{We further observe that the independent method tends to perform well when individual dimensions are independently accurate univariate classifiers and that it is credible that dependent approaches excel when there are strong mutual interdependencies between them with respect to the class.}
    
    
    \reviewone{
    Inspired by the Elastic Ensemble of nearest neighbor classifiers using different univariate distance measures, we then further experiment with ensembles of multivariate similarity \reviewtwo{and distance} measures and show that ensembling results in accuracy competitive with the state of the art.

    Our three ensembles establish a baseline in our future plans to create a multivariate TS-CHIEF which would combine similarity \reviewtwo{and distance}-based techniques with dictionary-based, interval-based for multivariate TSC.    
    }


\section*{Acknowledgment}

This research was supported by the Australian Research Council under grant DP210100072. This material is based upon work supported by the Air Force Office of Scientific Research, Asian Office of Aerospace Research and Development (AOARD) under award number FA2386-18-1-4030. 

The authors would like to thank Prof.~Anthony~Bagnall and everyone who contributed to the UEA multivariate time series classification archive.

    \bibliographystyle{IEEEtran}
    \bibliography{IEEEabrv,references}
    
   \clearpage

\appendix
\clearpage

\section{Summary of the Datasets}

\begin{table*}[ht]
\centering
\caption{Summary of the 23 fixed-length multivariate datasets we used from the UAE repository.}
\label{tbl:datasets}
\begin{tabular}{p{0.1cm}lp{0.6cm}*{3}{p{0.9cm}}*{3}{p{0.7cm}}}
\toprule
 \# &                    dataset &  code & trainsize & testsize &  dims & length & classes \\
\midrule
  1 &  ArticularyWordRecognition &   AWR &       275 &      300 &     9 &    144 &      25 \\
  2 &         AtrialFibrillation &    AF &        15 &       15 &     2 &    640 &       3 \\
  3 &               BasicMotions &    BM &        40 &       40 &     6 &    100 &       4 \\
  4 &                    Cricket &    CR &       108 &       72 &     6 &   1197 &      12 \\
  5 &              DuckDuckGeese &   DDG &        50 &       50 &  1345 &    270 &       5 \\
  6 &                   Epilepsy &    EP &       137 &      138 &     3 &    206 &       4 \\
  7 &       EthanolConcentration &    EC &       261 &      263 &     3 &   1751 &       4 \\
  8 &                      ERing &    ER &        30 &      270 &     4 &     65 &       6 \\
 9 &            FingerMovements &    FM &       316 &      100 &    28 &     50 &       2 \\
 10 &      HandMovementDirection &   HMD &       160 &       74 &    10 &    400 &       4 \\
 11 &                Handwriting &    HW &       150 &      850 &     3 &    152 &      26 \\
 12 &                  Heartbeat &    HB &       204 &      205 &    61 &    405 &       2 \\
 13 &                     Libras &   LIB &       180 &      180 &     2 &     45 &      15 \\
 14 &                       LSST &  LSST &      2459 &     2466 &     6 &     36 &      14 \\
 15 &                     NATOPS &  NATO &       180 &      180 &    24 &     51 &       6 \\
 16 &                  PenDigits &    PD &      7494 &     3498 &     2 &      8 &      10 \\
 17 &                    PEMS-SF &  PEMS &       267 &      173 &   963 &    144 &       7 \\
 18 &                    Phoneme &    PS &      3315 &     3353 &    11 &    217 &      39 \\
 19 &               RacketSports &    RS &       151 &      152 &     6 &     30 &       4 \\
 20 &         SelfRegulationSCP1 &  SRS1 &       268 &      293 &     6 &    896 &       2 \\
 21 &         SelfRegulationSCP2 &  SRS2 &       200 &      180 &     7 &   1152 &       2 \\
 22 &              StandWalkJump &   SWJ &        12 &       15 &     4 &   2500 &       3 \\
 23 &        UWaveGestureLibrary &    UW &       120 &      320 &     3 &    315 &       8 \\
\bottomrule
\end{tabular}
\end{table*}

\clearpage
\section{Accuracy of Dependent and Independent Measures}

\begin{table*}[ht]
\centering
\caption{Accuracy of independent similarity \reviewtwo{and distance} measures. Note that DTWF and DDTWF refers to measures that use full window.}
\label{tbl:acc-indep}
\begin{tabular}{p{0.6cm}*{11}{p{0.55cm}}}
\toprule
dataset &            L2 &          dtwf &         dtwcv &         ddtwf &        ddtwcv &          wdtw &         wddtw &          lcss &           msm &           erp &           twe \\
\midrule
    AWR &          0.98 & \textbf{0.99} &          0.99 &          0.62 &          0.71 &          0.99 &          0.72 &          0.99 &          0.98 &          0.99 &          0.97 \\
     AF &          0.31 &          0.21 &          0.25 &          0.23 &          0.30 &          0.29 &          0.33 &          0.29 &          0.29 &          0.33 & \textbf{0.33} \\
     BM &          0.57 & \textbf{1.00} &          1.00 &          0.99 &          1.00 &          1.00 &          0.99 &          0.91 & \textbf{1.00} &          0.92 & \textbf{1.00} \\
     CK &          0.90 & \textbf{1.00} & \textbf{1.00} &          0.97 &          0.97 &          1.00 &          0.96 &          0.98 &          0.99 &          0.97 &          0.98 \\
    DDG &          0.43 & \textbf{0.58} &          0.57 & \textbf{0.58} &          0.53 &          0.58 &          0.54 &          0.44 &          0.54 &          0.49 &          0.58 \\
     EP &          0.68 &          0.97 &          0.97 &          0.95 &          0.95 &          0.97 &          0.95 &          0.97 & \textbf{0.98} &          0.90 &          0.97 \\
     ER &          0.93 &          0.92 &          0.94 &          0.83 &          0.92 & \textbf{0.95} &          0.89 &          0.93 &          0.93 &          0.94 &          0.93 \\
     EC &          0.29 &          0.29 &          0.29 &          0.27 &          0.25 &          0.27 &          0.26 &          0.25 &          0.29 & \textbf{0.31} &          0.29 \\
     FM &          0.54 & \textbf{0.57} &          0.57 &          0.51 &          0.55 &          0.55 &          0.52 &          0.54 &          0.55 &          0.55 &          0.53 \\
    HMD &          0.28 &          0.30 &          0.31 &          0.32 &          0.31 &          0.32 &          0.30 &          0.31 &          0.31 &          0.24 & \textbf{0.33} \\
     HW &          0.31 &          0.48 &          0.48 &          0.27 &          0.29 & \textbf{0.48} &          0.29 &          0.45 &          0.47 &          0.38 &          0.35 \\
     HB &          0.66 &          0.68 &          0.69 &          0.71 & \textbf{0.71} &          0.69 &          0.70 & \textbf{0.71} &          0.69 &          0.66 &          0.70 \\
    LIB &          0.79 &          0.87 &          0.87 &          0.88 &          0.88 &          0.87 & \textbf{0.88} &          0.83 &          0.84 &          0.80 &          0.85 \\
   LSST &          0.45 & \textbf{0.57} &          0.56 &          0.48 &          0.48 &          0.57 &          0.48 &          0.29 &          0.54 &          0.45 &          0.52 \\
   NATO &          0.77 &          0.79 &          0.78 & \textbf{0.83} &          0.82 &          0.80 &          0.82 &          0.79 &          0.77 &          0.77 &          0.77 \\
   PEMS &          0.82 &          0.78 &          0.82 &          0.69 &          0.69 &          0.82 &          0.70 & \textbf{0.86} &          0.82 &          0.83 &          0.84 \\
     PD & \textbf{0.99} &          0.98 & \textbf{0.99} &          0.98 &          0.99 &          0.99 &          0.99 &          0.98 &          0.99 &          0.99 &          0.98 \\
     RS &          0.78 &          0.84 &          0.85 &          0.78 &          0.81 &          0.85 &          0.80 & \textbf{0.89} &          0.84 &          0.85 &          0.81 \\
   SRS1 &          0.80 &          0.80 & \textbf{0.81} &          0.57 &          0.56 &          0.81 &          0.56 &          0.80 &          0.80 &          0.80 &          0.77 \\
   SRS2 &          0.46 &          0.50 &          0.48 &          0.51 &          0.52 &          0.48 &          0.52 &          0.48 &          0.51 &          0.49 & \textbf{0.56} \\
    SWJ &          0.27 &          0.33 &          0.31 &          0.39 & \textbf{0.42} &          0.31 &          0.32 &          0.33 &          0.23 &          0.32 &          0.31 \\
     UW &          0.89 &          0.89 &          0.90 &          0.76 &          0.86 &          0.90 &          0.85 & \textbf{0.91} &          0.90 &          0.90 &          0.89 \\
     \bottomrule
   Wins &             1 &             6 &             3 &             2 &             2 &             2 &             1 &             4 &             2 &             1 &             4 \\
\bottomrule
\end{tabular}
\end{table*}

\begin{table}
\centering
\caption{Accuracy of dependent similarity \reviewtwo{and distance} measures. Note that dtwf and ddtwf refers to measures that use full window. Note that the accuracy for L2 differs in Table~\ref{tbl:acc-indep} and \ref{tbl:acc-dep} because instead of using the same $p$ (see Equations~\ref{eq:lp-indep} and~\ref{eq:lp-dep}), our experiments sum the distances for independent measures as proposed in~\cite{shokoohi2017generalizing}.}
\label{tbl:acc-dep}
\begin{tabular}{p{0.6cm}*{11}{p{0.55cm}}}
\toprule
dataset &            L2 &          dtwf &         dtwcv &         ddtwf &        ddtwcv &          wdtw &         wddtw & lcss &           msm &           erp &           twe \\
\midrule
    AWR &          0.97 & \textbf{0.99} &          0.99 &          0.34 &          0.34 &          0.98 &          0.34 & 0.98 &          0.98 &          0.98 &          0.97 \\
     AF & \textbf{0.35} &          0.21 &          0.31 &          0.23 &          0.29 &          0.34 &          0.30 & 0.34 &          0.33 &          0.29 &          0.33 \\
     BM &          0.58 & \textbf{0.96} & \textbf{0.96} &          0.89 &          0.89 & \textbf{0.96} &          0.89 & 0.76 &          0.76 &          0.74 &          0.89 \\
     CK &          0.92 & \textbf{1.00} & \textbf{1.00} &          0.82 &          0.83 & \textbf{1.00} &          0.82 & 0.99 &          1.00 &          0.98 &          0.96 \\
    DDG &          0.42 & \textbf{0.48} &          0.48 &          0.35 &          0.35 &          0.47 &          0.35 & 0.40 &          0.29 &          0.38 &          0.25 \\
     EP &          0.65 & \textbf{0.96} &          0.95 &          0.90 &          0.89 &          0.95 &          0.89 & 0.93 &          0.94 &          0.89 &          0.92 \\
     ER &          0.93 &          0.94 & \textbf{0.94} &          0.82 &          0.85 &          0.93 &          0.84 & 0.92 &          0.91 &          0.93 &          0.92 \\
     EC &          0.29 &          0.30 &          0.29 &          0.24 &          0.24 &          0.29 &          0.25 & 0.31 & \textbf{0.32} &          0.29 &          0.26 \\
     FM &          0.55 & \textbf{0.55} &          0.54 &          0.50 &          0.50 &          0.55 &          0.50 & 0.49 &          0.52 & \textbf{0.55} &          0.53 \\
    HMD &          0.27 &          0.31 & \textbf{0.33} &          0.29 &          0.26 &          0.31 &          0.30 & 0.22 &          0.21 &          0.23 &          0.25 \\
     HW &          0.31 & \textbf{0.61} &          0.61 &          0.40 &          0.39 &          0.61 &          0.40 & 0.54 &          0.57 &          0.46 &          0.46 \\
     HB &          0.63 &          0.70 &          0.68 & \textbf{0.71} &          0.70 &          0.68 &          0.71 & 0.64 &          0.66 &          0.70 &          0.71 \\
    LIB &          0.79 &          0.88 &          0.87 &          0.93 &          0.93 &          0.88 & \textbf{0.93} & 0.32 &          0.86 &          0.80 &          0.87 \\
   LSST &          0.45 &          0.55 &          0.55 &          0.43 &          0.43 & \textbf{0.55} &          0.43 & 0.36 &          0.36 &          0.45 &          0.44 \\
   NATO &          0.79 &          0.83 &          0.82 & \textbf{0.84} &          0.84 &          0.82 &          0.83 & 0.72 &          0.80 &          0.79 &          0.80 \\
   PEMS &          0.78 &          0.77 &          0.78 &          0.60 &          0.60 &          0.78 &          0.60 & 0.14 &          0.78 &          0.78 & \textbf{0.78} \\
     PD &          0.99 &          0.99 &          0.99 &          0.99 &          0.99 & \textbf{0.99} &          0.99 & 0.98 &          0.97 &          0.99 &          0.99 \\
     RS &          0.79 &          0.85 &          0.87 &          0.80 &          0.83 &          0.87 &          0.81 & 0.87 & \textbf{0.88} &          0.77 &          0.83 \\
   SRS1 &          0.80 & \textbf{0.82} &          0.82 &          0.54 &          0.53 &          0.81 &          0.53 & 0.77 &          0.78 &          0.80 &          0.53 \\
   SRS2 &          0.46 &          0.53 &          0.52 &          0.48 &          0.51 &          0.51 &          0.50 & 0.47 &          0.52 &          0.49 & \textbf{0.55} \\
    SWJ &          0.28 &          0.21 &          0.31 &          0.27 & \textbf{0.36} &          0.33 &          0.33 & 0.35 &          0.26 &          0.29 &          0.29 \\
     UW &          0.88 &          0.92 &          0.92 &          0.84 &          0.85 & \textbf{0.92} &          0.86 & 0.91 &          0.91 &          0.88 &          0.88 \\
        \bottomrule
   Wins &             1 &             8 &             4 &             2 &             1 &             5 &             1 &    0 &             2 &             1 &             2 \\
\bottomrule
\end{tabular}
\end{table}

\end{document}